\documentclass[a4paper,fleqn]{cas-dc}

\usepackage[numbers]{natbib}
\usepackage{float}

\newcommand{\mb}[1]{\mathbf{#1}}

\def\tsc#1{\csdef{#1}{\textsc{\lowercase{#1}}\xspace}}
\tsc{WGM}
\tsc{QE}

\begin{document}
\let\WriteBookmarks\relax
\def\floatpagepagefraction{1}
\def\textpagefraction{.001}

\shorttitle{End-to-end Neural Network Based Quadcopter control}

\shortauthors{R. Ferede et~al.}

\title [mode = title]{End-to-end Neural Network Based Optimal Quadcopter Control}                      

\tnotetext[1]{This work was supported by ESA}


%
\author[1]{Robin Ferede}[orcid=0000-0001-5158-565X]

\cormark[1]


\ead{r.ferede@tudelft.nl}



\affiliation[1]{organization={Micro Air Vehicle lab, Control and Simulation, Delft University of Technology},
    addressline={Kluyverweg 1}, 
    city={Delft},
    postcode={2629 HS}, 
    country={The Netherlands}}

\author[1]{Guido {de Croon}}[orcid=0000-0001-8265-1496]
\ead{G.C.H.E.deCroon@tudelft.nl}

\author[1]{Christophe {De Wagter}}[orcid=0000-0002-6795-8454]
\ead{C.deWagter@tudelft.nl}


\affiliation[2]{organization={Advanced Concepts Team, European Space Agency},
    addressline={Keplerlaan 1}, 
    city={Noordwijk},
    postcode={2201 AZ}, 
    country={The Netherlands}}

\author[2]{Dario Izzo}[orcid=0000-0002-9846-8423]
\ead{dario.izzo@esa.int}


\cortext[cor1]{Corresponding author}



\begin{abstract}
Developing optimal controllers for aggressive high-speed quadcopter flight poses significant challenges in robotics. Recent trends in the field involve utilizing neural network controllers trained through supervised or reinforcement learning. However, the sim-to-real transfer introduces a reality gap, requiring the use of robust inner loop controllers during real flights, which limits the network's control authority and flight performance. In this paper, we investigate for the first time, an end-to-end neural network controller, addressing the reality gap issue without being restricted by an inner-loop controller. The networks, referred to as G\&CNets, are trained to learn an energy-optimal policy mapping the quadcopter's state to rpm commands using an optimal trajectory dataset. In hover-to-hover flights, we identified the unmodeled moments as a significant contributor to the reality gap. To mitigate this, we propose an adaptive control strategy that works by learning from optimal trajectories of a system affected by constant external pitch, roll and yaw moments. In real test flights, this model mismatch is estimated onboard and fed to the network to obtain the optimal rpm command. We demonstrate the effectiveness of our method by performing energy-optimal hover-to-hover flights with and without moment feedback. Finally, we compare the adaptive controller to a state-of-the-art differential-flatness-based controller in a consecutive waypoint flight and demonstrate the advantages of our method in terms of energy optimality and robustness.
\end{abstract}


\begin{highlights}
\item First flight tested, end-to-end neural network controller for quadcopters that does not rely on inner loop controllers
\item Successful hover-to-hover flight revealed a significant contribution of unmodeled moments to the reality gap
\item Proposed adaptive control strategy to learn from optimal trajectories of perturbed systems.
\item Adaptive network automatically finds new optimal trajectories for perturbed systems
\end{highlights}


\maketitle

\section{Introduction}
Nowadays there is an increasing demand for autonomous quadcopters for various military and civilian applications \cite{Hassanalian2017ClassificationsAA}. For many applications such as emergency response, inspection, delivery or racing the drone must fly as fast, and as energy efficient as possible \cite{POFlight}. However, developing autonomous systems for aggressive high-speed flight still poses many challenges. One of these challenges is developing computationally efficient optimal control algorithms that take into account non-linear dynamics and actuator limits.

Current state-of-the-art research on time-optimal quadcopter control focuses on making controllers track a reference guidance trajectory. Popular tracking methods include the  differential-flatness-based controller (DFBC) \cite{NIEUWSTADT19962301, MinSnap, Faessler2018, tal2020accurate} and the traditional nonlinear-model-predictive controller (NMPC) \cite{aerospace4020031, Bicego2020, explicitMPC, torrente2021data, MPCC, TimeOpimalReplanning}. While the DFBC is more computationally efficient, traditional NMPC has gained a lot of popularity in quadcopter control due to advances in hardware. The advantages of NMPC over DFBC are improved tracking accuracy for dynamically infeasible trajectories as well as improved robustness to model mismatch \cite{MPC_DFBP} (especially by means of adaptive algorithms \cite{AdaptiveNMPC, torrente2021data}). Furthermore, in recent work, a traditional NMPC method was shown to outperform human pilots in a drone-racing task by tracking offline-generated time-optimal trajectories \cite{TimeOptimalPlanning}. 
\begin{figure*}
    \centering
    \includegraphics[width=\linewidth]{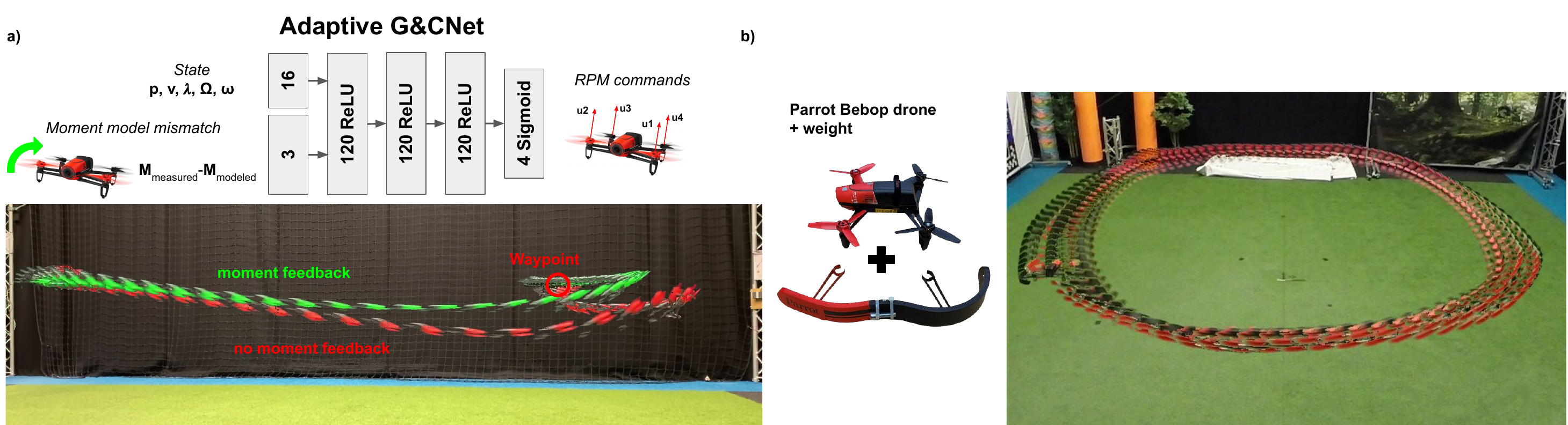}
    \caption{a) The reality gap issue is resolved by estimating the moment model mismatch and feeding it to the adaptive G\&CNet.  b) We perturb the dynamics by adding a weight on one side, the Bebop drone successfully flies through the 3$\times$4m track by adapting its rpm command based on the observed state and moment model mismatch.}
    \label{fig:fig1}
\end{figure*}

An inherent limitation with all of these methods is that the aggressiveness and efficiency of the performed maneuver are fully determined by the trajectory to be tracked. Moreover, the generation of time-optimal trajectories is often computationally intensive and requires either offline calculation, or an online sub-optimal simplification in the form of polynomial guidance \cite{MinSnap, Mueller, tal2020accurate}, point mass trajectories \cite{TimeOpimalReplanning, tankasala2022smooth} or numerical approximation methods \cite{numericaltrajgen, successive_convexification, yu2022real}. Additionally, to add control authority to the algorithms, a margin is often defined to lower the actuator limits used for the trajectory generation. This reduces optimality since time-optimal control relies on saturating the actuators in a bang-bang fashion. Furthermore in order to improve robustness, the algorithms are never given direct motor control in real-life flights. Instead, the algorithm sends higher-level commands such as thrust and rates to an inner-loop controller.

A recent trend in quadcopter control research is the application of machine learning techniques to trajectory generation and tracking. Deep neural networks have been trained for trajectory generation using reinforcement learning \cite{Autonomous_Drone_Racing_with_Deep_Reinforcement_Learning} and  supervised machine learning \cite{LearnTraj}. Similarly, trajectory tracking has been improved by training neural networks either from flight data \cite{DNN_improved_tracking, RBFNN} or from simulation data \cite{RL_wp_stab, deep_drone_acrobatics}. Another research line \cite{DBLP:journals/corr/abs-1912-07067, LandingProblems, Tailor} proposes an alternative to the trajectory tracking-based control methods by combining guidance and control into a single neural network (termed G\&CNet) which is trained to imitate the optimal state feedback from a dataset of time-optimal trajectories. Once trained, the G\&CNet provides a computationally efficient way to compute the optimal control onboard the quadcopter without requiring any trajectory (re)planning. With a real flight test, this has been demonstrated to work for longitudinal trajectories based on a simplified, 2-dimensional quadcopter model \cite{DBLP:journals/corr/abs-1912-07067}. In these experiments, the G\&CNet was used to calculate thrust and pitch acceleration commands which were tracked by an INDI\cite{Smeur2016} controller.

In this article, we take the G\&CNet approach a step further and investigate for the first time an end-to-end, i.e., state-to-rpm network for a 3-dimensional quadcopter model taking into account drag, aerodynamic effects and actuator delays. Unlike the previous work \cite{DBLP:journals/corr/abs-1912-07067} this network can fully exploit the 6-degrees-of freedom of the quadcopter model. Furthermore, our network directly calculates the rpm motor commands which allows us to take advantage of the actuator's limits without being limited by a low-level controller. The biggest obstacle with this approach is the reality gap between the model and the real world. In this research, we identify this reality gap for energy-optimal flight and propose an adaptive method to mitigate the effects of unmodeled roll, pitch and yaw moments. Furthermore, we benchmark our controller's performance against a state-of-the-art differential-flatness-based controller using an identical setup with the same hardware. Here we demonstrate the advantages of our method in terms of energy optimality and robustness.

\section{Methodology}
\subsection{Quadcopter model}
\begin{figure}
    \centering
    \includegraphics[width=0.6\linewidth]{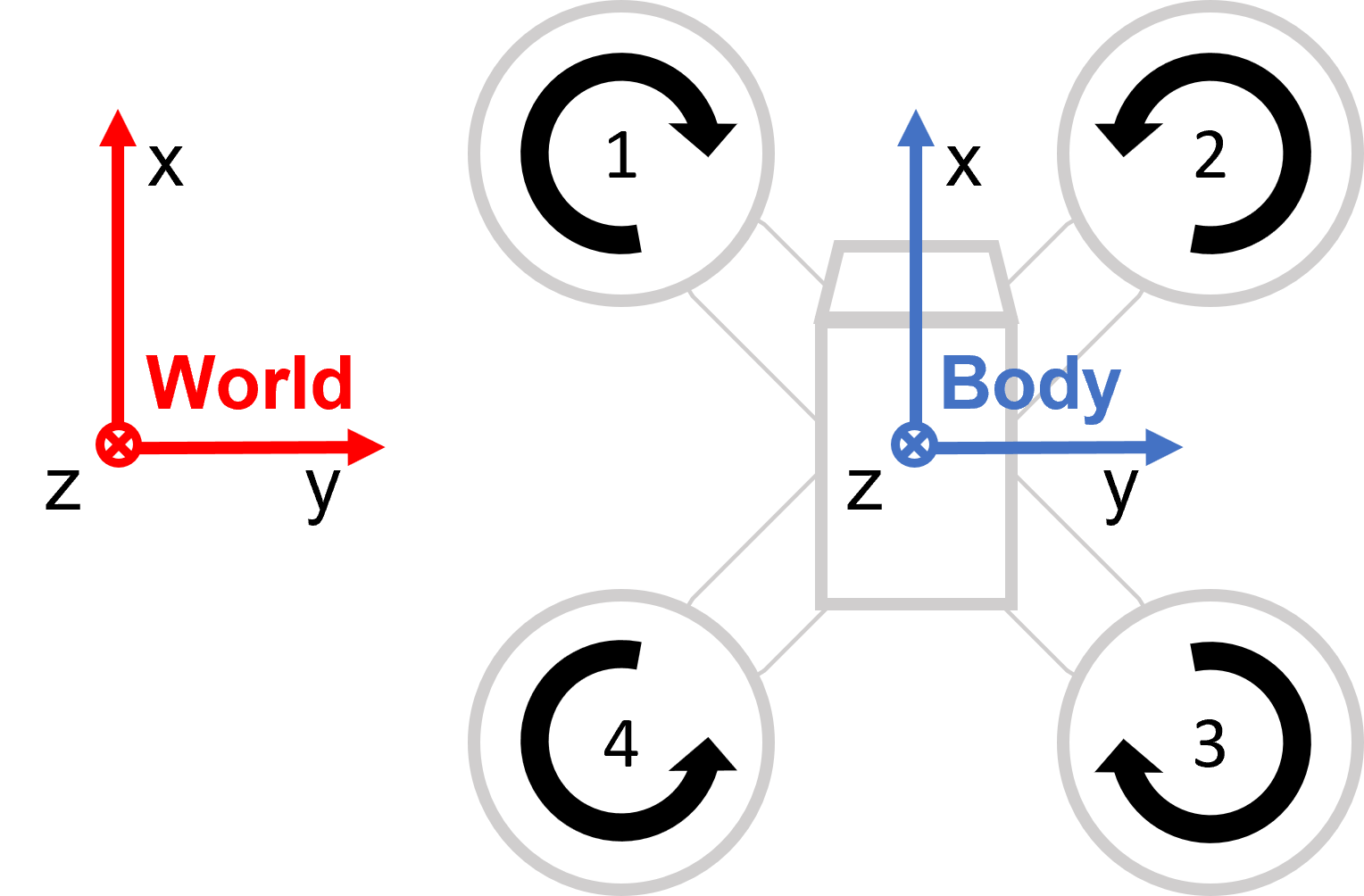}
    \caption{Quadcopter configuration and axes definition (z-axis points downwards)}
    \label{fig:axis_def}
\end{figure}

Referring to the quadcopter configuration and axes definition illustrated in Figure\ref{fig:axis_def}, the state and control input of the quadcopter can be described as follows:
\begin{align*}
    \mb{x} = [\mb{p}, \mb{v}, \boldsymbol \lambda, \mb{\Omega}, \mb{\boldsymbol \omega}]^T \quad \mb{u} = [u_1, u_2, u_3, u_4]^T
\end{align*}
Where $\mb{p} = [x, y, z]$ and $\mb{v} = [v_x, v_y, v_z]$ are the position and velocity in the world frame, $\mb{\Omega} = [p, q, r]$ is the angular velocity in body frame, $\boldsymbol \lambda = [\phi, \theta, \psi]$ are the Euler angles that describe the orientation of the body frame and $\boldsymbol \omega = [\omega_1, \omega_2, \omega_3, \omega_4]$ are the angular velocities of each of the propellers in rpm. The control input $\mb{u}$ contains the normalized rpm commands $u_i \in [0,1]$. The system dynamics are described by:
\begin{align} \label{eq:quadcopter_model} 
    \dot{\mb{p}} &= \mb{v} &
    \dot{\mb{v}} &= \mb{g} + R(\mb{\lambda}) \mb{F}\\
    \dot{\lambda} &= Q(\mb{\lambda}) \mb{\Omega} &
    I \dot{\mb{\Omega}} &= - \mb{\Omega} \times I \mb{\Omega} + \mb{M} \nonumber \\
    &&\dot{\boldsymbol \omega}  &= ((\omega_{max}-\omega_{min})\mb{u}+\omega_{min} - \boldsymbol \omega)/\tau  \nonumber
\end{align}
Where $\mb{g}= [0, 0, g]^T$ is the gravitational acceleration, $I$ is the moment of inertia matrix given by $\text{diag}(I_x,I_y,I_z)$, $\omega_{min}$ and $\omega_{max}$ are the minimum and maximum propeller rpm limits and $\tau$ is the first order delay parameter of the actuator model. Furthermore, $R(\mb{\lambda})$ is the rotation matrix defined by:
\begin{align*}
    R(\mb{\lambda})
    &=
    \begin{bmatrix}
        c_{\theta}c_{\psi} & -c_{\phi}s_{\psi}+s_{\phi}s_{\theta}c_{\psi} & s_{\phi}s_{\psi}+c_{\phi}s_{\theta}c_{\psi} \\
        c_{\theta}s_{\psi} & c_{\phi}c_{\psi}+s_{\phi}s_{\theta}s_{\psi} & -s_{\phi}c_{\psi}+c_{\phi}s_{\theta}s_{\psi} \\
        -s_{\theta} & s_{\phi}c_{\theta} & c_{\phi}c_{\theta}
    \end{bmatrix}
\end{align*}
and $Q(\mb{\lambda})$ denotes a transformation between angular velocities and Euler angles. $\mb{F} = [F_x, F_y, F_z]^T$ is the specific force acting on the quadcopter in the body frame which we model as a function of the body velocities and the propeller RPMs using a thrust and drag model based on \cite{Svacha2017ImprovingQT}:
\begin{align} \label{eq:F_model}
\begin{split}
    F_x &= - k_x v^B_x \sum_{i=1}^4 \omega_i \quad F_y = - k_y v^B_y \sum_{i=1}^4 \omega_i\\
    F_z &= -k_\omega \sum_{i=1}^4 \omega_i^2 - k_z v^B_z \sum_{i=1}^4 \omega_i - k_h (v^{B2}_x + v^{B2}_y)
\end{split}
\end{align}
Similarly, $\mb{M}=[M_x, M_y, M_z]^T$ is the moment acting on the quadcopter which we model with the following equations:
\begin{align} \label{eq:M_model}
\begin{split}
    M_x &= k_p (\omega_1^2 - \omega_2^2 - \omega_3^2 + \omega_4^2) + k_{pv} v^{B}_y\\
    M_y &= k_q (\omega_1^2 + \omega_2^2 - \omega_3^2 - \omega_4^2) + k_{qv} v^{B}_x\\
    M_z &= k_{r1} (-\omega_1 + \omega_2 - \omega_3 + \omega_4) \\
    & + k_{r2} (-\dot{\omega_1} + \dot{\omega_2} - \dot{\omega_3} + \dot{\omega_4})  - k_{rr} r\\
\end{split}
\end{align}
See Table\ref{tab:model_parameters} for the parameter values identified for our platform.

\begin{table*}
    \centering
    \caption{Model parameters for the Parrot Bebop quadcopter. The moments of inertia $I_x, I_y, I_z$ are obtained from \cite{Sun2019}. All other parameters have been identified by means of linear regression with sensor data obtained from various flights}
    \label{tab:model_parameters}
    \begin{tabular}{c|c|c|c|c|c|c|c}
    \hline
    \hline
    $k_x$ $_{[\text{rpm}^{-1} \text{s}^{-1}]}$ & $k_y$ $_{[\text{rpm}^{-1} \text{s}^{-1}]}$ & $k_\omega$ $_{[\text{rpm}^{-2} \text{m} \text{s}^{-2}]}$ & $k_z$ $_{[\text{rpm}^{-1} \text{s}^{-1}]}$ & $k_h$ $_{[m^{-1}]}$ & $I_x$ $_{[\text{kg} \text{m}^2]}$ & $I_y$ $_{[\text{kg} \text{m}^2]}$ & $I_z$ $_{[\text{kg} \text{m}^2]}$\\
    \hline
    1.08e-05 & 9.65e-06 & 4.36e-08 & 2.79e-05 & 6.26e-02 &  0.000906 &  0.001242 & 0.002054\\
    \hline
    \hline
    $k_p$ $_{[\text{rpm}^{-2} \text{N} \text{m}]}$ & $k_{pv}$ $_{[\text{N} \text{s}]}$ & $k_q$ $_{[\text{rpm}^{-2} \text{N} \text{m}]}$ & $k_{qv}$ $_{[\text{N} \text{s}]}$ & $k_{r1}$ $_{[\text{rpm}^{-1} \text{N} \text{m}]}$ & $k_{r2}$ $_{[\text{rpm}^{-1} \text{N} \text{m} \text{s}]}$ & $k_{rr}$ $_{[\text{N} \text{m} \text{s}]}$  & $\tau$ $_{[\text{s}]}$ \\
    \hline
    1.41e-09 & -7.97e-03 & 1.22e-09 & 1.29e-02 & 2.57e-06 & 4.11e-07 & 8.13e-04 & 0.06 \\
    \hline
    \hline
    \end{tabular}
\end{table*}

\subsection{Energy optimal control problem}
Given a state space $X$ and set of admissible controls $U$, the goal is to find a control trajectory $\mb{u} : [0,T] \to U$ that steers the system from an initial state $\mathbf{x}_0$ to some target state $S \subset X$ in time $T$ while minimizing some cost function. The energy optimal control problem considered in this paper is formulated as
\begin{align} \label{eq:OCP}
\begin{split}
    \underset{\mb{u}, T}{\text{minimize}} \quad &E(\mb{u}, T) =  \int_{0}^{T} ||\mb{u}(t)||^2 dt\\
    \text{subject to} \quad &\dot{\mb{x}} = f(\mb{x}, \mb{u}) \quad \mb{x}(0) = \mb{x}_0 \quad \mb{x}(T) \in S
\end{split}
\end{align}
Similar to \cite{DBLP:journals/corr/abs-1912-07067} the control problem is transformed into a Nonlinear Programming (NLP) problem using Hermite Simpson transcription. The trajectories $\mb{x}(t), \mb{u}(t)$ are discretized into $N+1$ points with a time step $\Delta t = T/N$ such that $\mb{x}_k = \mb{x}(k \Delta t)$ and $\mb{u}_k = \mb{u}(k \Delta t)$ Using the AMPL \cite{fourer1990modeling} modeling language with the SNOPT NLP solver \cite{gill2005snopt}, the optimal (discretized) trajectory $\mb{x}^*_0 \ldots \mb{x}_N^*$ and $\mb{u}^*_0 \ldots \mb{u}_N^*$ can be computed.\\

\subsection{Dataset generation and network training}
A dataset is created by generating optimal trajectories for a range of initial conditions. From these trajectories, a dataset of state-action pairs can be obtained of the form $(\mb{x}_i^*, \mb{u}_i^*) \quad i = 0, \ldots, N$. We use these state-action pairs to train a Neural Network $f_N: X \to U$ to approximate the optimal feedback\footnote{From \cite{LandingProblems}:  "the Hamilton-Jacobi-Bellman equations are important here as they imply the existence and uniqueness of an optimal state-feedback $\mb{u}^*(\mb{x})$ which, in turn, allow to consider universal function approximators such as deep neural networks to represent it."} that maps $\mb{x}_i^*$ to $\mb{u}_i^*$. In all our experiments we use a  neural network with 3 hidden layers of 120 neurons with ReLU activation and an output layer of 4 neurons with Sigmoid activation (Fig \ref{fig:fig1}). Similar to \cite{DBLP:journals/corr/abs-1912-07067} we use the mean squared error loss function:
\begin{align*}
    l=|| f_N(\mb{x}^*_i) - \mb{u}_i^* ||^2
\end{align*}
with mini-batch size 256 and a starting learning rate of 1e-3.

\subsection{Adaptive Method} \label{sec:adaptive_method}
We modify our model by assuming the existence of some constant external moment $\mb{M}_{ext} = [M_{ext, x}, M_{ext, y}, M_{ext, z}]^T$ acting on the system. The external moment can thus be considered part of our state vector $\mb{x} = [\mb{p}, \mb{v}, \mb{\lambda}, \Omega, \mb{\omega}, \mb{M}_{ext}]^T$ The modified system dynamics becomes:
\begin{align} \label{eq:quadcopter_model_modified} 
    \dot{\mb{p}} &= \mb{v} &
    \dot{\mb{v}} &= \mb{g} + R(\mb{\lambda}) \mb{F}\\
    \dot{\lambda} &= Q(\mb{\lambda}) \mb{\Omega} &
    I \dot{\mb{\Omega}} &= - \mb{\Omega} \times I \mb{\Omega} + \mb{M} + \mb{M}_{ext} \nonumber \\
    \dot{\mb{M}}_{ext} &= 0 &\dot{\boldsymbol \omega}  &= ((\omega_{max}-\omega_{min})\mb{u}+\omega_{min} - \boldsymbol \omega)/\tau  \nonumber
\end{align}
Using the same approach as before, we can now generate optimal trajectories for this system and train a network to approximate the optimal state feedback. Additionally, the neural network will now have 3 extra inputs for $M_{ext, x}, M_{ext, y}, M_{ext, z}$. The obtained controller will now use these extra inputs to optimally compensate for the unmodeled moments (assuming they are constant). For the onboard implementation, we will obtain the values of $\mb{M}_{ext}$ by subtracting the modeled moment (Eq. ~\ref{eq:M_model}) from the measured moment
\begin{align} \label{eq:M_measured}
    M_{measured} = I\dot{\mb{\Omega}}+\mb{\Omega}\times I \mb{\Omega}
\end{align}
using filtered (8Hz 2nd order Butterworth low-pass filter) gyroscope measurements. It is important to note that the filtering causes our estimates for $\mb{M}_{ext}$ to be slightly delayed. Furthermore, the controller's output is based on the assumption of a constant external moment so we can expect our method to only be effective if the modeling errors are in a sufficiently low-frequency range.

\subsection{Differential-flatness-based Controller (DFBC)} \label{sec:MinSnap}
DFBC is a state-of-the-art method for generating aggressive trajectories using piece-wise high-order polynomials $\mb{p}(t) = [x(t), y(t), z(t), \psi(t)]^T$ that pass through a set of waypoints while minimizing the 'Snap' defined by the following integral \cite{MinSnap}:
\begin{align*}
    \int_0^T \mu_r \Big[ x^{(4)}(t) + y^{(4)}(t)+ z^{(4)}(t) \Big]^2 + \mu_{\psi} \Big[ \psi^{(2)}(t)^2 \Big] dt
\end{align*}
In this problem, the final time is fixed, and the polynomial coefficients are found by solving a quadratic constraint optimization problem. As shown in \cite{MinSnap}, if we change the final time by a factor of $\alpha$, the new minimum snap solution is simply a time-scaled version of the original polynomial $\mb{p}(\alpha t)$. By changing the value of $\alpha$, the trajectory can be faster or slower without having to recompute the optimal solution. In order to achieve accurate tracking, we use an outer-loop INDI controller where the velocity and acceleration feed-forward commands are directly computed from the polynomials.
\section{Experimental Setup}
The quadcopter used in our experiment is the Parrot Bebop~1 which has its onboard software replaced by the Paparazzi-UAV open-source autopilot project \cite{Gati2013}. All computations will run in real time on the Parrot P7 dual-core CPU Cortex A9 processor. The Parrot Bebop has an MPU6050 IMU sensor that will be used to obtain measurements of the specific force and angular velocity along the body axes. Additionally, the Bebop can measure the angular velocities (in rpm) of each of the propellers, which is a requirement for our control method.

\begin{figure*}
    \centering
    \includegraphics[width=\linewidth]{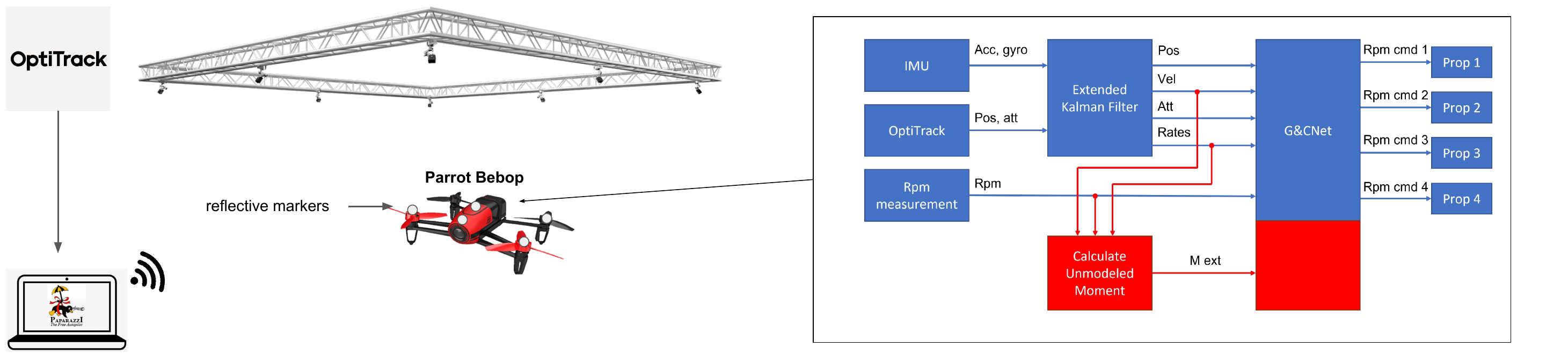}
    \caption{Experimental setup: The Parrot Bebop's position and attitude are tracked with OptiTrack and sent via WiFi, while an onboard extended Kalman filter fuses the OptiTrack and IMU data to get an accurate state estimate for the G\&CNet.}
    \label{fig:experimental_setup}
\end{figure*}

All flight tests are performed in The CyberZoo which is a research and test laboratory in the faculty of Aerospace Engineering at the TU Delft. This lab consists of a 10 by 10 meter area surrounded by nets with an OptiTrack motion capture system that can provide position and attitude data in real-time. An extended Kalman filter is used to fuse the OptiTrack and IMU data to obtain an estimate of the position, velocity, attitude and body rates. These state variables are used as input to the G\&CNet along with the rpm measurements. The outputs of the network will be directly used as rpm commands to the propellers. The DFBC method will use the same state estimates to obtain the feedforward terms for the INDI controller. See Figure\ref{fig:experimental_setup} for an overview of the experimental setup.
\section{Results \& Discussion}
\subsection{Nominal G\&CNet}
\subsubsection{Dataset and network}
Using the system dynamics from equation \ref{eq:quadcopter_model} we generate a dataset of 100,000 energy-optimal trajectories with a target hover state defined by $\mb{x}, \mb{v}, \mb{\lambda}, \mb{\Omega}, \dot{\mb{v}}, \dot{\mb{\Omega}}, \dot{\boldsymbol \omega} = 0$. The rpm limits are set to $\omega_{min} = 5000, \quad \omega_{max}=10000$ and the initial conditions are uniformly sampled from the following intervals:
\begin{align*}
    x &\in [-5,5] &
    y &\in [-5,5] &
    z &\in [-1,1] \\
    v_x &\in [-\tfrac{1}{2},\tfrac{1}{2}] &
    v_y &\in [-\tfrac{1}{2},\tfrac{1}{2}] &
    v_z &\in [-\tfrac{1}{2},\tfrac{1}{2}] \\
    \phi &\in [-\tfrac{2 \pi}{9},\tfrac{2 \pi}{9}] &
    \theta &\in [-\tfrac{2 \pi}{9},\tfrac{2 \pi}{9}] &
    \psi &\in [-\pi,\pi] \\
    p &\in [-1,1] &
    q &\in [-1,1] &
    r &\in [-1,1] \\
    \boldsymbol \omega &\in [\omega_{min},\omega_{max}]^4 &&&
\end{align*}
We split this dataset into a training set of 90,000 trajectories and a test set of 10,000 trajectories. The G\&CNet is trained until a mean squared error of $\sim$0.0003 is obtained on the test set.
\subsubsection{Simulation and flight test}
\begin{figure*}
    \centering
    \includegraphics[width=\linewidth]{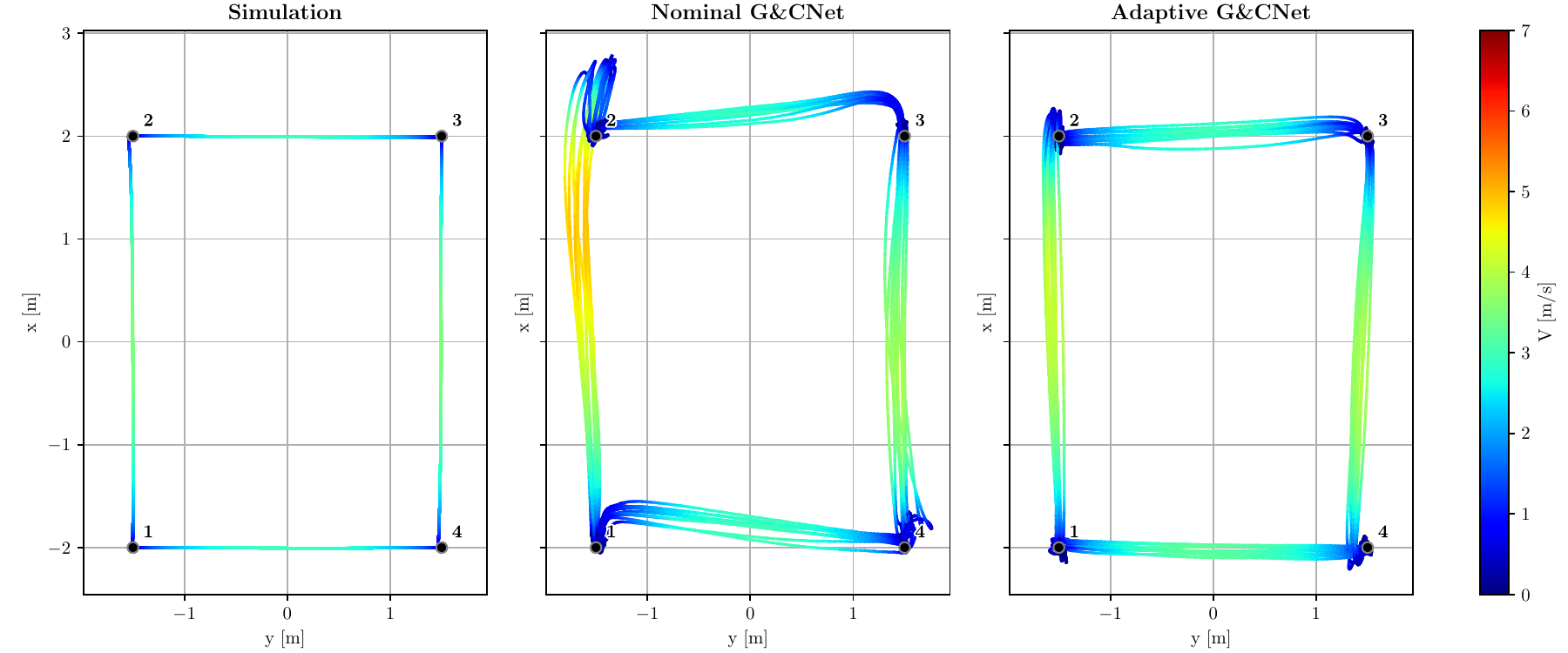}
    \caption{Top-down view of the simulated trajectory next to the Nominal- and Adaptive G\&CNet flight test}
    \label{fig:top_view_sim_nom_ada}
\end{figure*}
\begin{figure}
    \centering
    \includegraphics[width=\linewidth]{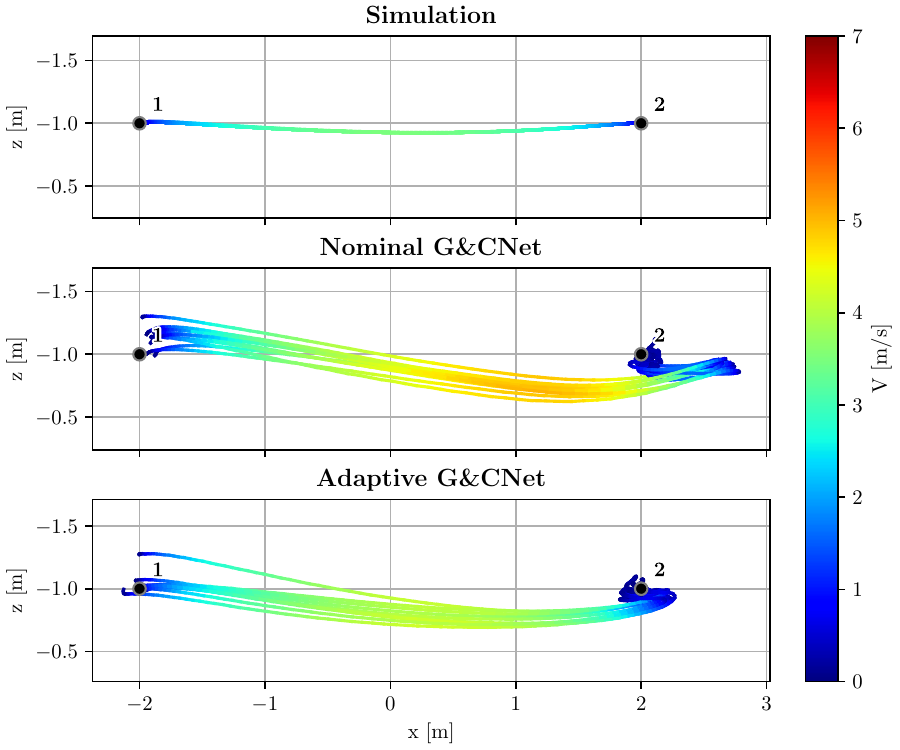}
    \caption{Sideways view of the trajectories between waypoints 1 and 2: Simulation next to the Nominal- and Adaptive G\&CNet flight test. \vspace{-5mm}}
    \label{fig:sim_vs_flight_forward}
\end{figure}
With the trained nominal G\&CNet, we simulate the closed loop system dynamics and do a flight test where the drone flies from hover to hover in a 3$\times$4m rectangle. In order to fly to the target waypoints, we subtract the waypoint coordinates from the $x,y$ and $z$ neural network inputs. Both in simulation and the flight test, the drone flies 10 laps in which the target waypoint is switched every 4 seconds. In Figure\ref{fig:top_view_sim_nom_ada} a top-down view of the trajectory can be seen for the simulation and the flight test. As expected, in the simulation, the trajectories show significant overlap and the drone consistently arrives at the waypoint without overshooting. In the flight test, the trajectories are more spread out and a large deviation can be seen in the positive x-direction. The unmodeled effects are especially visible in the forward translation maneuver where the drone speeds up too much and overshoots the next waypoint. In Figure\ref{fig:sim_vs_flight_forward} these forward trajectories are shown from a sideways view. It can be seen that the drone loses too much altitude causing it to speed up and overshoot.
\subsubsection{Unmodeled Effects}
\begin{figure}
     \centering
    \includegraphics[width=\linewidth]{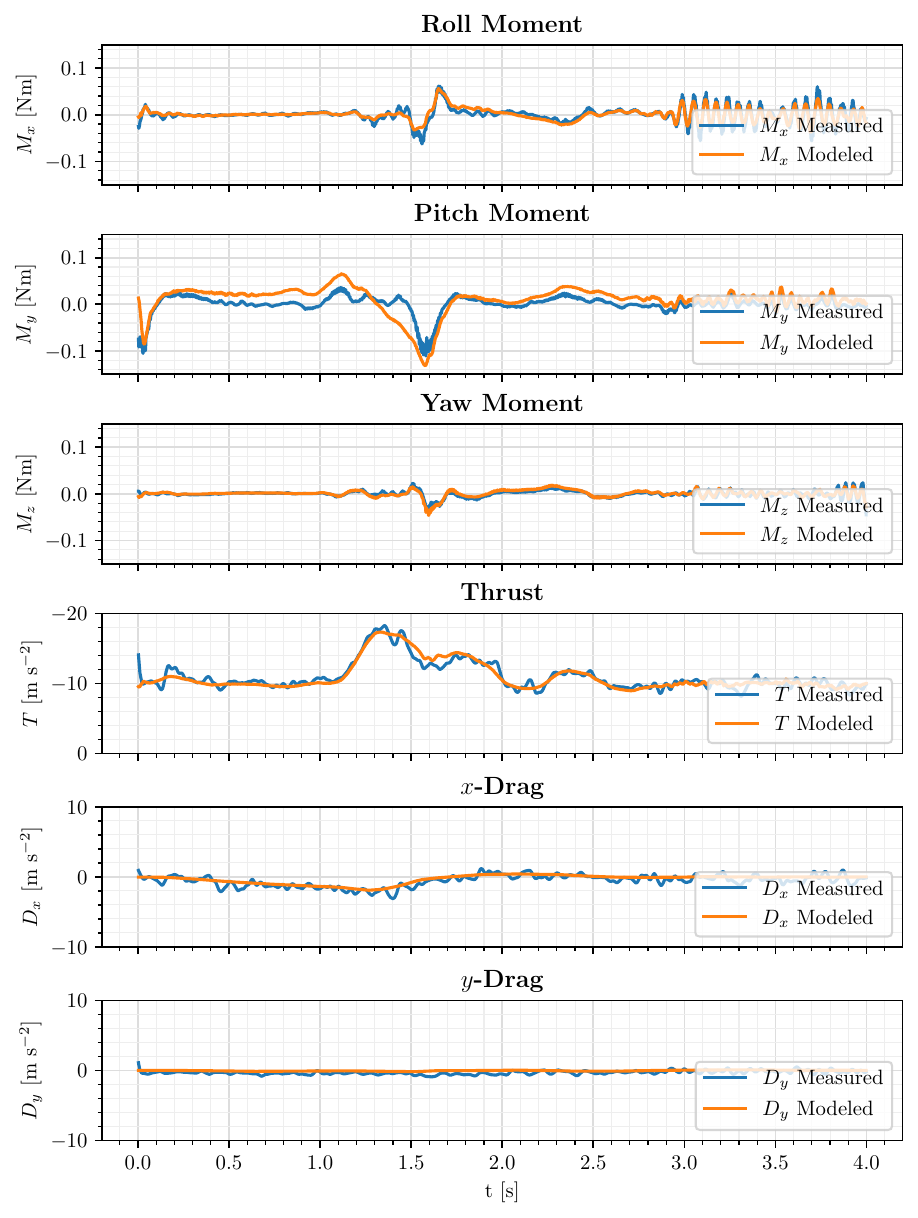}
     \caption{Comparison of the measured and modeled moments and (specific) forces encountered in one of 'Nominal G\&CNet' flights from Figure\ref{fig:sim_vs_flight_forward}}
     \vspace{-1mm}
     \label{fig:measured_modeled}
 \end{figure}

We investigate the unmodeled aerodynamic effects from the forward translation flight by comparing the measured and modeled moments and specific forces. The measured moments and forces are obtained by using the filtered (16Hz 2nd order Butterworth non-causal filter) gyroscope and accelerometer measurements. Figure\ref{fig:measured_modeled} shows these measured and modeled quantities for one of the forward translation trajectories of the nominal G\&CNets from Figure\ref{fig:sim_vs_flight_forward}.
It can be observed that the pitch moment seems to have a significant low-frequency model mismatch. The unmodeled pitch moment is mostly negative which might explain why the drone is diving down so much in the flight test. Because our current parametric model cannot capture this effect, we choose to go for an adaptive control strategy.

\subsection{Adaptive G\&CNet}
\subsubsection{Dataset and network}
We use the modified system dynamics with external moments from equation \ref{eq:quadcopter_model_modified} to generate another 100,000 energy-optimal trajectories with the same target state and initial conditions as before, only now we also uniformly sample the external moments from the following intervals:
\begin{align*}
    M_{x,ext},M_{y,ext} &\in [-0.04, 0.04] &
    M_{z,ext} &\in [-0.01, 0.01]
\end{align*}
With the generated dataset we train the adaptive G\&CNet with 3 extra $M_{ext}$ inputs to learn the optimal state feedback for the modified system. Again, we train until a mean squared error of $\sim$0.0003 is achieved.

\subsubsection{Performance comparison}
\begin{figure}
    \centering
    \includegraphics[width=\linewidth]{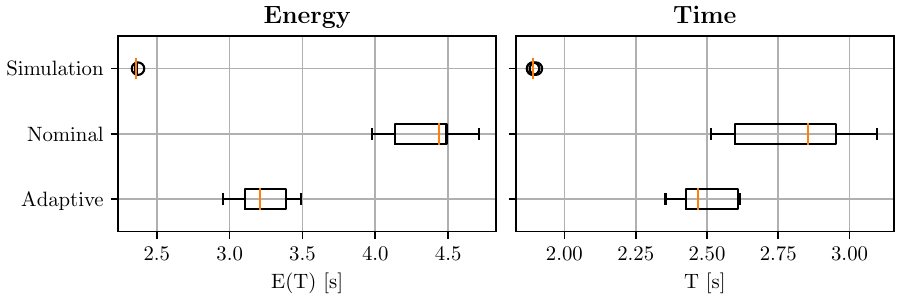}
    \caption{Energy and time comparison during the 4m forward flight between waypoints 1 and 2: Simulation compared to Nominal and Adaptive G\&CNet. \vspace{-5mm}}
    \label{fig:HoverPT}
\end{figure}

With the adaptive G\&CNet, we perform the same flight test using the 4 waypoints and compare the results to the nominal network. In Figure\ref{fig:top_view_sim_nom_ada} and \ref{fig:sim_vs_flight_forward} the trajectory is compared to the previous nominal network and the simulation. It can be seen that the trajectory no longer deviates towards the positive x-direction and the overshoot in the forward translation maneuver is significantly reduced. Furthermore the box-plot in Figure\ref{fig:HoverPT} shows the arrival time $T$ and energy $E(T)=\int_0^T || \mb{u}(t) ||^2 dt$ corresponding to the trajectories from Figure\ref{fig:sim_vs_flight_forward}. As one might expect, the performance gain of the adaptive network is most significant in terms of Energy. However, the arrival time and energy in the flight tests are still significantly higher than in simulation which is probably due to the remaining unmodeled effects causing the overshoot at the 2nd waypoint
\subsection{Bench-marking: Adaptive G\&CNet vs. DFBC} \label{sec:benchmarking}
\subsubsection{Adaptive G\&CNet}
For the task of flying through consecutive waypoints, we will train an adaptive G\&CNet to reach the waypoint with a forward final velocity in the direction of a $45^\circ$ yaw angle. Using the modified system dynamics from Eq.~\ref{eq:quadcopter_model_modified} we generate a dataset of 10,000 energy-optimal trajectories with a target state given by:
\begin{align*}
    x,y,z,v_z,p,q,r,\dot{p},\dot{q},\dot{r} = 0, \tfrac{v_y}{v_x} = \tan (\tfrac{\pi}{4}), \psi = \tfrac{\pi}{4}
\end{align*}
The rpm limits are set to $\omega_{min} = 3000$, $\omega_{max}=12000$
and the initial conditions are uniformly sampled from the following intervals:
\begin{align*}
    x &\in [-5,-2] &
    y &\in [-1,1] &
    z &\in [-\tfrac{1}{2},\tfrac{1}{2}] \\
    v_x &\in [-\tfrac{1}{2},5] &
    v_y &\in [-3,3] &
    v_z &\in [-1,1] \\
    \phi &\in [-\tfrac{2 \pi}{9},\tfrac{2 \pi}{9}] &
    \theta &\in [-\tfrac{2 \pi}{9},\tfrac{2 \pi}{9}] &
    \psi &\in [-\tfrac{\pi}{3},\tfrac{\pi}{3}] \\
    p &\in [-1,1] &
    q &\in [-1,1] &
    r &\in [-1,1] \\
    \boldsymbol \omega &\in [\omega_{min},\omega_{max}]^4 
\end{align*}
We split this dataset into a training set of 9000 trajectories and a test set of 1000 trajectories and train until a mean squared error of $\sim$0.0003 is obtained on the test set. With the trained adaptive G\&CNet we perform a flight test where we fly through 4 waypoints in a 3$\times4$m rectangle (See Figure\ref{fig:PO_benchmark} and \ref{fig:TimeBenchmarkWeight}). The controller switches to the next target waypoint and changes the coordinate system once the drone is within 1.2m from the current target. When switching to the next waypoint, we rotate our coordinate system by $90^\circ$ (around the z-axis) and set the next waypoint as the origin.

\begin{figure}[ht]
    \centering
    \includegraphics[width=\linewidth]{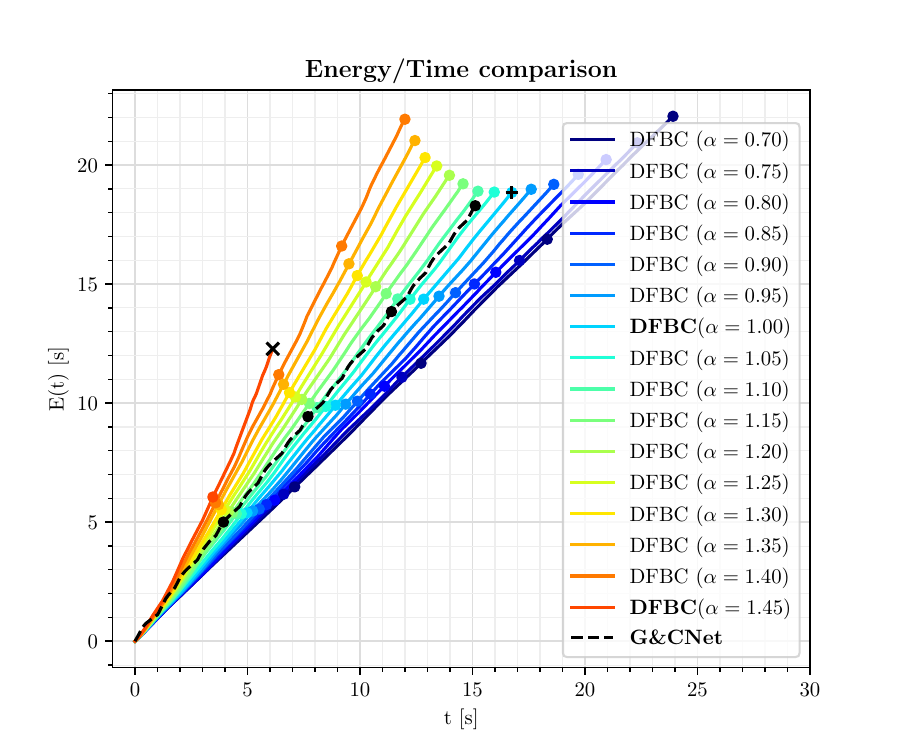}
    \caption{Energy plotted over time  during 4 laps of the $3 \times 4$m track. The points in time where a lap is completed are represented by a dot. The DFBC method that uses the least energy is marked with a "$+$". The flight that crashes is marked with an "$\times$". \vspace{-5mm}}
    \label{fig:BenchmarkPT}
\end{figure}

\subsubsection{DFBC}
We generate a piece-wise 6th order polynomial $\mb{p}(t) = [x(t), y(t), z(t), \psi(t)]^T$ that passes through 10 laps of the 4x3m track with a final time of 40 seconds. To make sure the trajectory starts in hover, the initial velocity, acceleration and yaw of the trajectory are set to 0. At the 2nd waypoint, we constrain the yaw angle to be $45^\circ$ which we increment by $90^\circ$ for each of the following waypoints. Additionally, at these waypoints, we constrain the velocity to be aligned with the yaw direction. Using the time scaling values starting at $\alpha=0.7$ we generate faster and faster trajectories by incrementing $\alpha$ by $0.05$. We then track these trajectories with the INDI controller for 4 laps. We increased alpha until the INDI controller could no longer track the trajectory (resulting in a crash). An overview of all the performed flights can be found in Figure\ref{fig:MinSnapAll} and \ref{fig:MinSnapWeightAll} in the appendix.
\subsubsection{Energy/Time comparison} \label{sec:energy_time}
\begin{figure}
    \centering    \includegraphics[width=\linewidth]{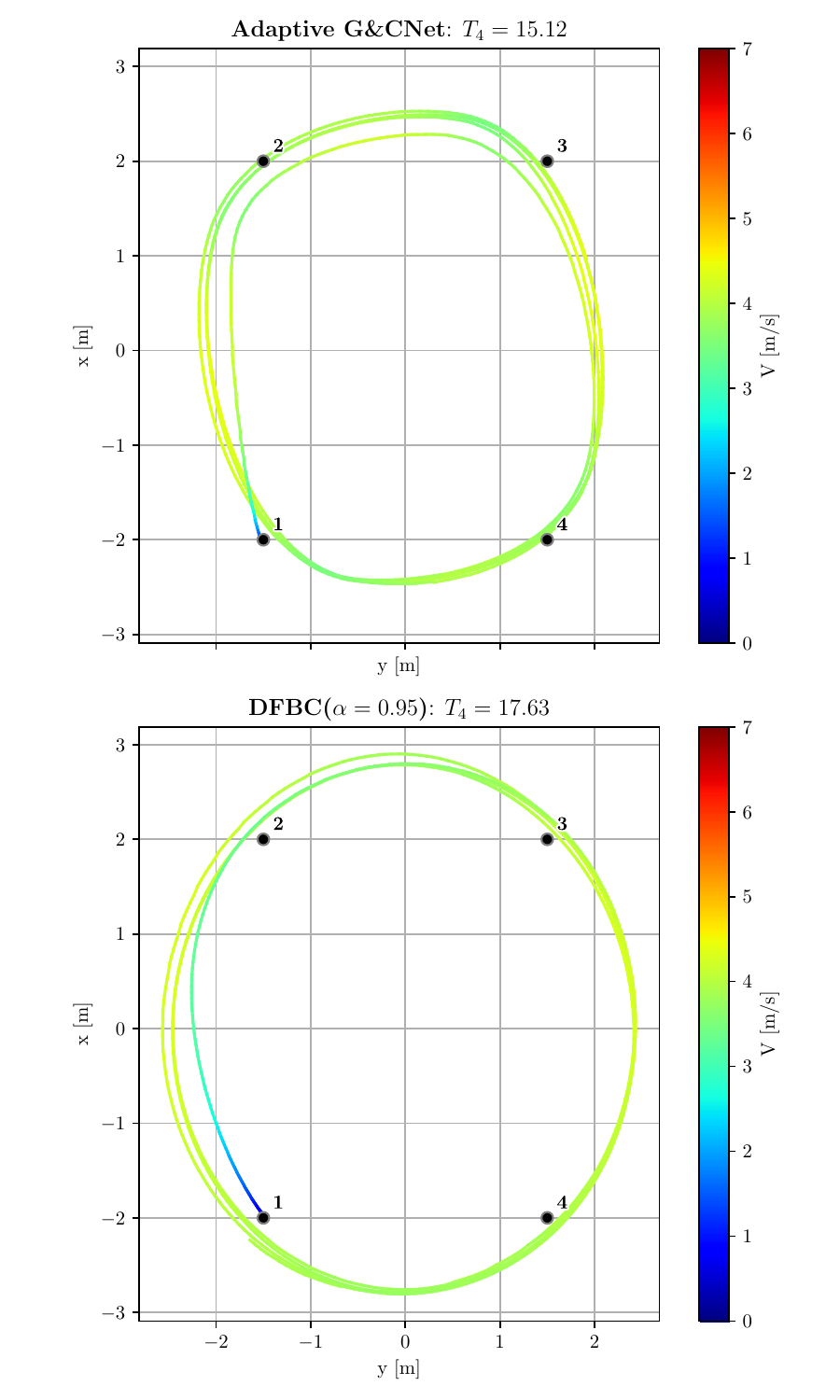}
    \vspace{-5mm}
    \caption{Top-down view of the adaptive G\&CNet's flight and the 'energy optimal' DFBC's flight at $\alpha=1.0$.}
    \label{fig:PO_benchmark}
\end{figure}

We now compare the lap times and the energy integral obtained from the flight tests. In Figure\ref{fig:BenchmarkPT} the energy integral $E(t) = \int_0^t ||\mb{u}(\tau)||^2 d\tau$ is plotted over time for the adaptive G\&CNet flight and all DFBC flights.
It can be noted that the fastest DFBC method finishes the 4 laps significantly faster than the G\&CNet. In terms of energy, however, the adaptive G\&CNet outperforms all of the DFBC methods. The DFBC method that uses the least energy ($\alpha = 1.0$) still uses more energy and time to finish the track. In Figure\ref{fig:PO_benchmark}, a top-down view of the trajectory of the 'energy optimal' DFBC method is plotted next to the adaptive G\&CNets flight. It can be seen that the DFBC method travels in a smooth circular trajectory at a relatively high velocity, while the G\&CNet takes tighter corners and flies at a lower velocity while still finishing the 4 laps quicker.
\subsubsection{Robustness experiment} \label{sec:robustness_experiment}
\begin{figure}
    \centering    \includegraphics[width=0.5\linewidth]{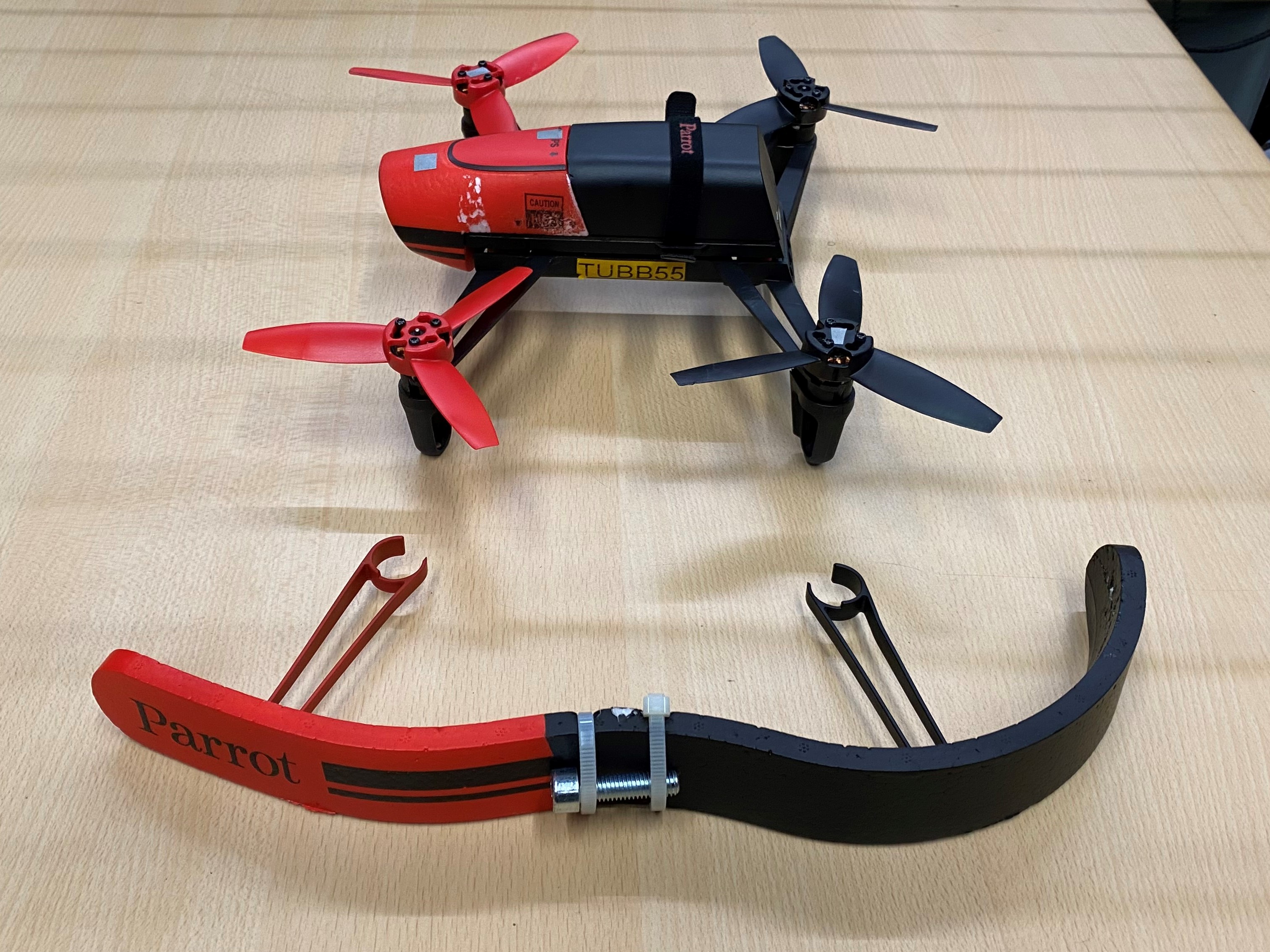}
    \caption{The Parrot Bebop drone and its bumper with a weight added to it. This imbalanced weight causes a roll moment of ~-0.06Nm  \vspace{-5mm}}
    \label{fig:Bebop+weight}
\end{figure}

\begin{figure}
    \centering
    \includegraphics[width=\linewidth]{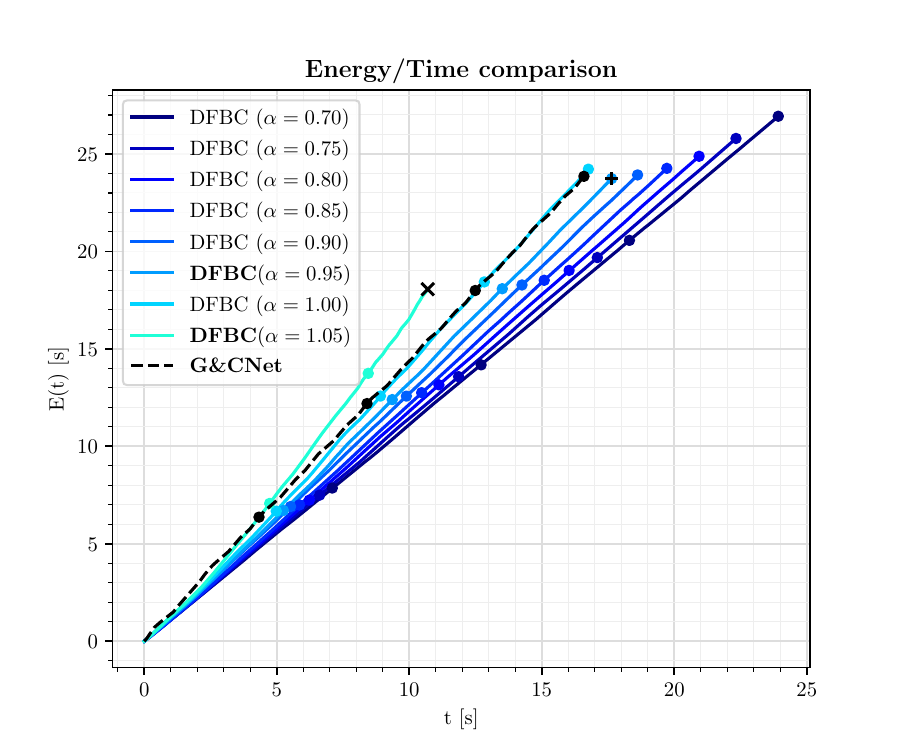}
    \vspace{-6mm}
    \caption{Energy plotted over time  during 4 laps of the $3 \times 4$m track with the added weight. The points in time where a lap is completed are represented by a dot. The DFBC method that uses the least energy is marked with a "$+$". The flight that crashes is marked with an "$\times$". \vspace{-5mm}}
    \label{fig:BenchmarkPTweight}
\end{figure}

\begin{figure}
    \centering
    \includegraphics[width=\linewidth]{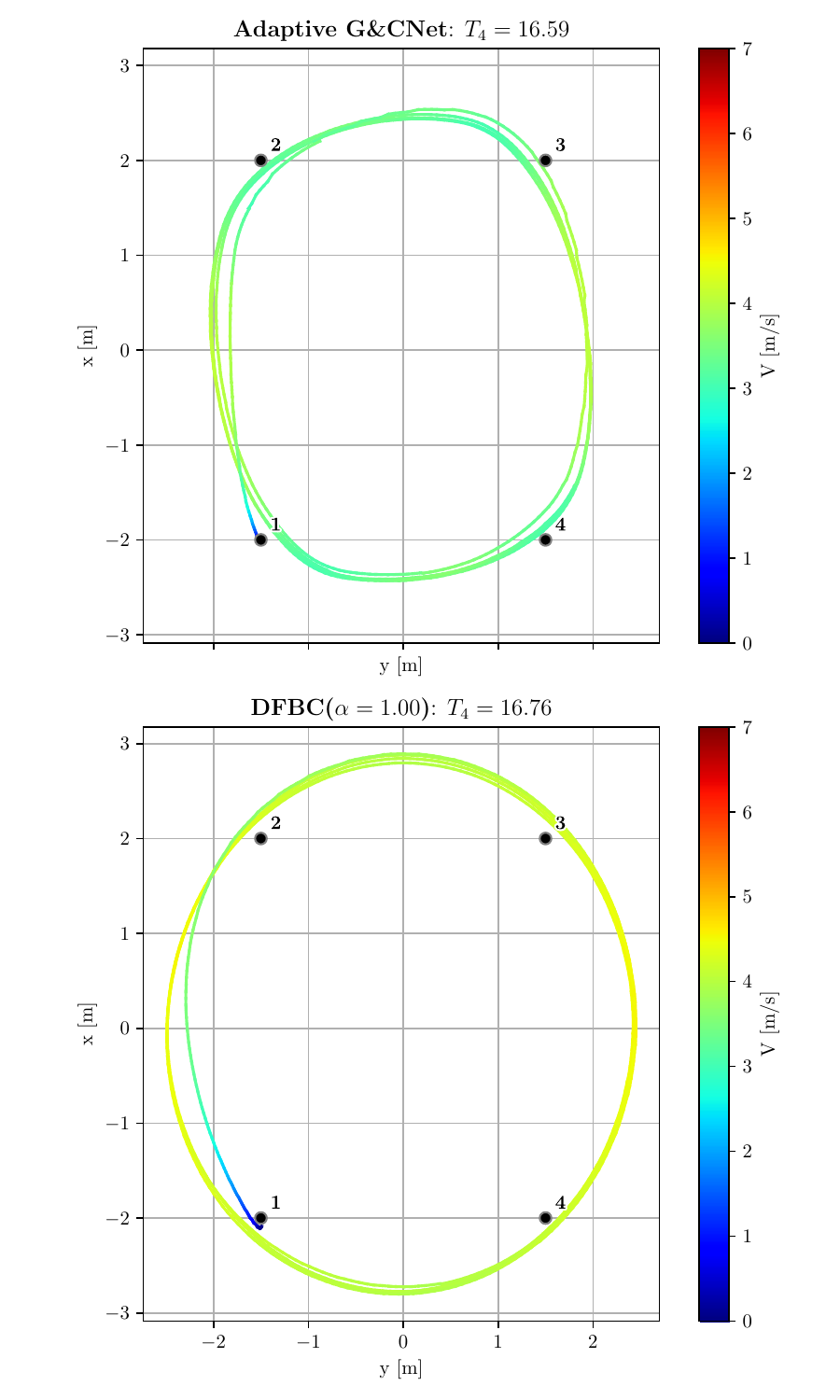}
    \vspace{-6mm}
    \caption{Top-down view of the adaptive G\&CNet's flight and the fastest DFBC's flight at $\alpha=1.0$ with the added weight. 
    \vspace{-5mm}}
    \label{fig:TimeBenchmarkWeight}
\end{figure}

\begin{figure}
    \centering
    \includegraphics[width=\linewidth]{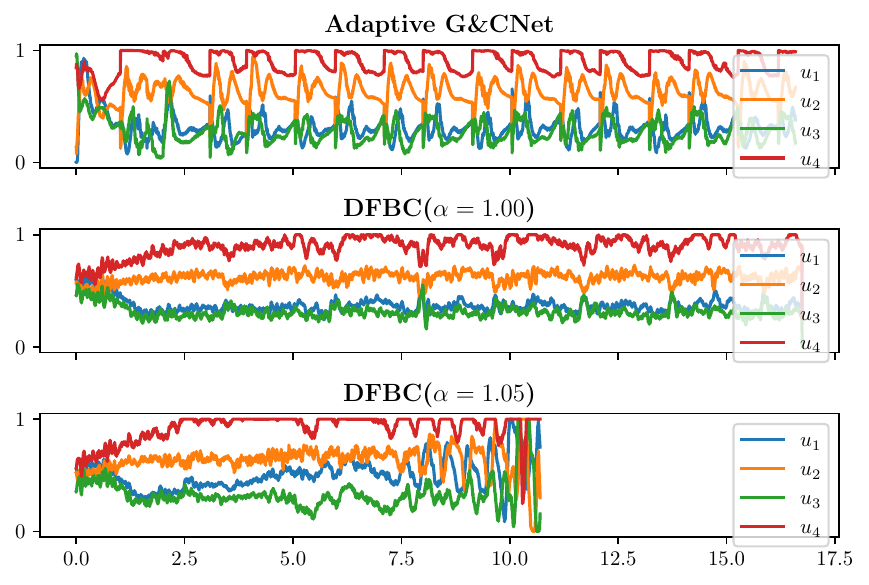}
    \vspace{-3mm}
    \caption{Comparison of the normalized RPM commands of the Adaptive G\&CNet compared to the fastest DFBC flight at $\alpha=1.00$ and the failed flight at $\alpha=1.05$.}
    \vspace{-1mm}
    \label{fig:RPMcomparison}
\end{figure}

In order to compare robustness, we apply an external moment to the drone by adding a bumper with a weight on the left side of the Bebop (Fig. \ref{fig:Bebop+weight}). With this alteration, we perform the same flight tests as before. In Figure\ref{fig:BenchmarkPTweight} we again show the energy/time plot for all of the performed flights. It can be seen that the DFBC controller fails a lot earlier at $\alpha=1.05$. In terms of time, the adaptive G\&CNet demonstrated superior performance, with the quadcopter flying faster than all of the DFBC flights. The trajectories of the G\&CNet and the fastest DFBC flight can be seen in Figure\ref{fig:TimeBenchmarkWeight}. Another interesting observation is that the G\&CNet flies slower with this added weight than it did in the previous flight. Here our method exhibits a clear advantage over DFBC, as it doesn't require a reference trajectory, and can dynamically adjust its course in real time. Furthermore, if we compare the rpm commands of both methods (Fig. \ref{fig:RPMcomparison}) it can be seen that the G\&CNet can handle sustained rpm saturations, while the DFBC method at $\alpha=1.05$ experiences similar saturations (at the same propeller) and crashes.

\section{Conclusion}
We have presented a novel G\&CNet setup to perform energy-optimal end-to-end control for a 3-dimensional quadcopter model (Eq. \ref{eq:quadcopter_model}). With real flights, we have investigated the performance of this G\&CNet, revealing that unmodeled moments were negatively influencing flight performance. To mitigate these effects, we proposed and implemented an adaptive control strategy that shows a significant improvement in flight performance. Furthermore, we compare our proposed adaptive G\&CNet to a DFBC method in consecutive waypoint flight scenarios, revealing clear advantages of our method over DFBC. Specifically, our method is more energy efficient, robust against large disturbances, and more flexible, with the ability to dynamically adjust its path in real time without relying on a reference trajectory.

Future work can focus on making current G\&CNets time-optimal while retaining their current robustness. This could be achieved by not only compensating for the unmodeled moments but also errors in thrust, drag forces and actuator delay. Additionally, robustness could be increased by using less strict final state constraints in the optimal control problem. Finally, to improve the maneuverability of the quadcopter in turns, the G\&CNet can be trained with optimal trajectories that account for two or more consecutive waypoints.

\section*{Acknowledgment}
This research was co-funded under the Discovery programme of, and funded by, the European Space Agency

\appendix
\section{Varying Altitude}
\begin{figure}[ht]
    \centering
    \includegraphics[width=\linewidth]{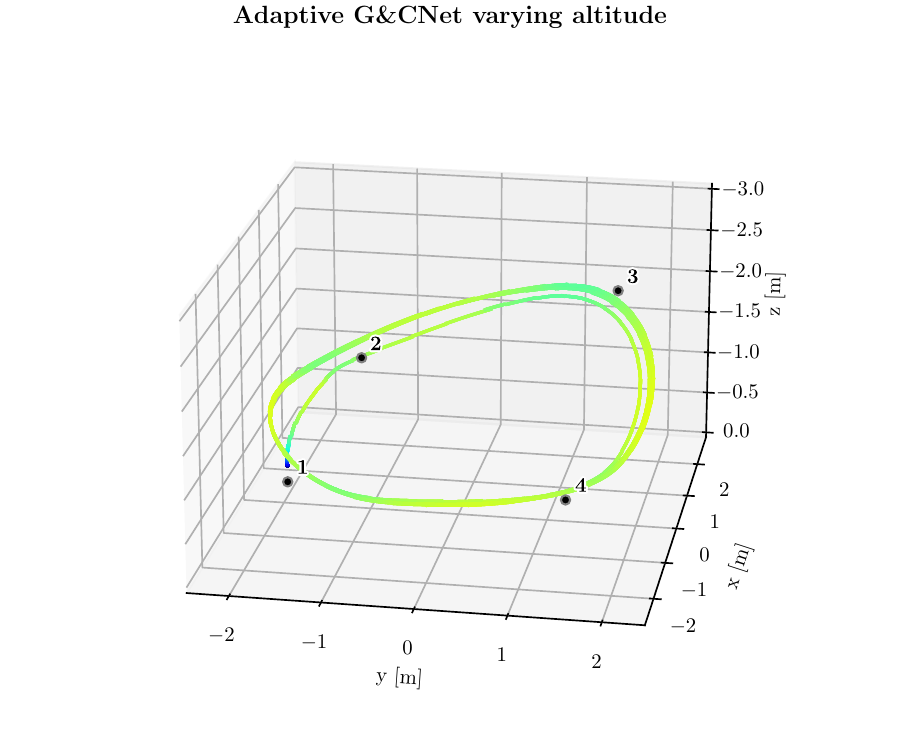}
    \vspace{-5mm}
    \caption{Trajectory of the adaptive G\&CNet through a 4$\times$3m track where the 3rd waypoint is raised by 1 meter.}
    \vspace{-5mm}
    \label{fig:GCNetAlt}
\end{figure}

The adaptive G\&CNet utilized in section \ref{sec:benchmarking} has, thus far, only been used to fly through a set of waypoints constrained to a horizontal plane. To exhibit the versatility of the trained G\&CNet, we will now execute a flight along the same 3$\times$4m track, but with one waypoint positioned 1 meter higher in altitude. Figure\ref{fig:GCNetAlt} shows the trajectory of this flight. Remarkably, even though the network was trained with a narrow range of +-0.5m in $z$ variation, it adeptly navigates through all the waypoints. This demonstration not only underscores the network's capability to navigate complex 3D paths but also its ability to generalize to some degree beyond the provided dataset.

\section{DFBC trajectories}
\begin{figure*}[ht]
    \centering
    \includegraphics[width=\linewidth]{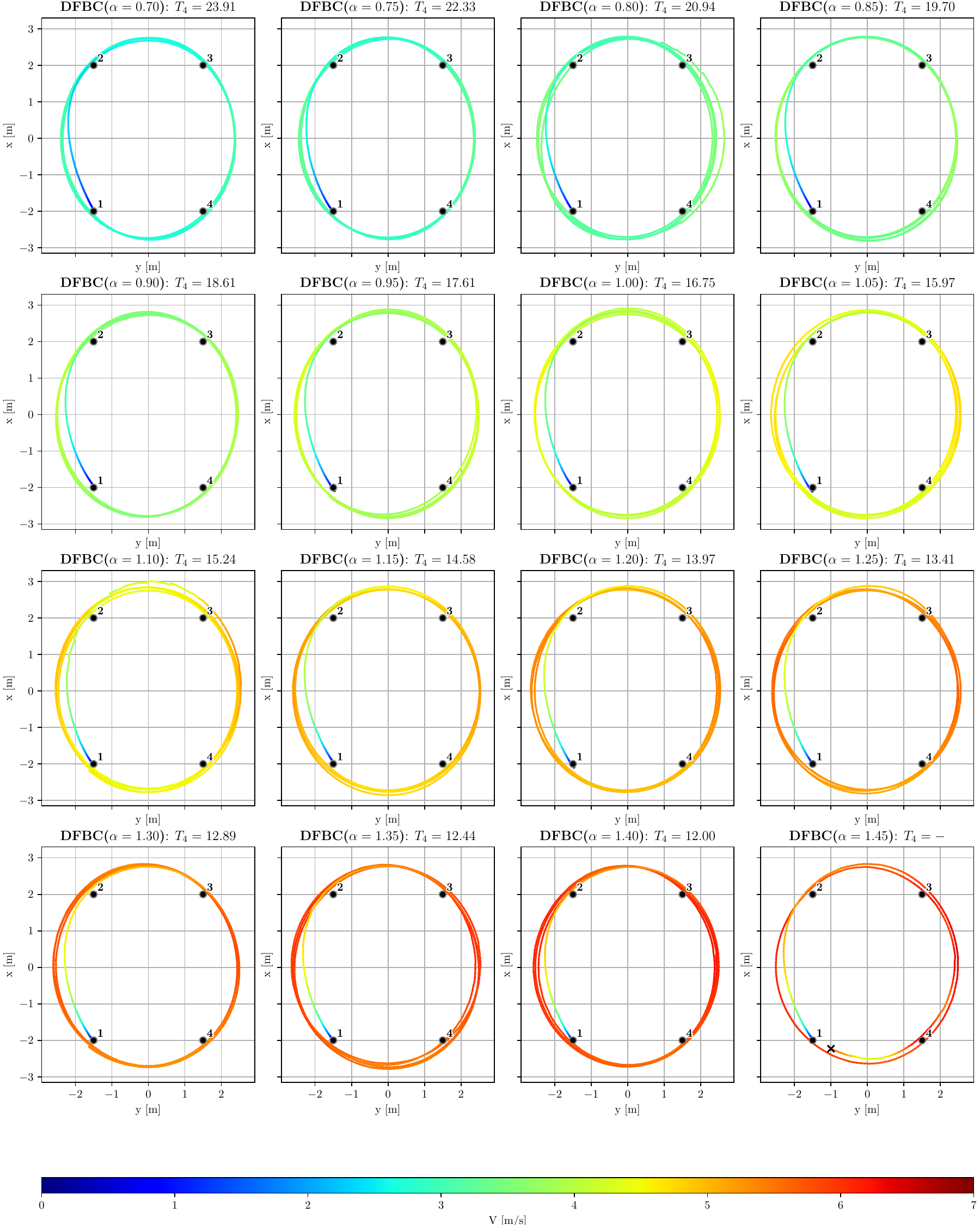}
    \caption{Top down view of all the DFBC trajectories used in the energy/time comparison from Section \ref{sec:energy_time}}
    \vspace{-0mm}
    \label{fig:MinSnapAll}
\end{figure*}

\begin{figure*}[ht]
    \centering
    \includegraphics[width=\linewidth]{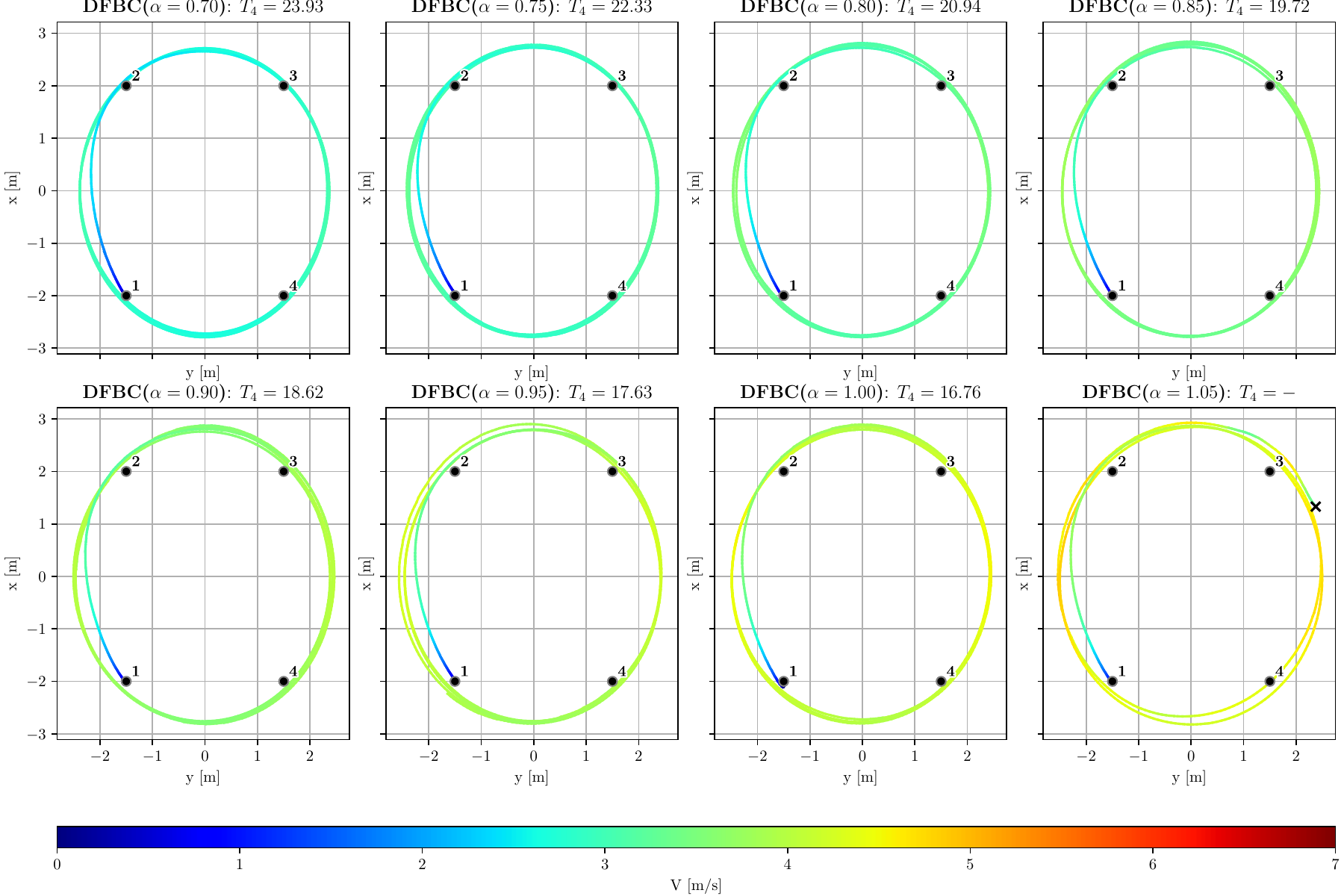}
    \caption{Top down view of all the DFBC trajectories used in the robustness experiment from Section \ref{sec:robustness_experiment}}
    \vspace{-4mm}
    \label{fig:MinSnapWeightAll}
\end{figure*}

Figure\ref{fig:MinSnapAll} and \ref{fig:MinSnapWeightAll} show a top-down view of all the performed DFBC flights.

\bibliographystyle{cas-model2-names}

\bibliography{root}

\begin{thebibliography}{35}
\expandafter\ifx\csname natexlab\endcsname\relax\def\natexlab#1{#1}\fi
\providecommand{\url}[1]{\texttt{#1}}
\providecommand{\href}[2]{#2}
\providecommand{\path}[1]{#1}
\providecommand{\DOIprefix}{doi:}
\providecommand{\ArXivprefix}{arXiv:}
\providecommand{\URLprefix}{URL: }
\providecommand{\Pubmedprefix}{pmid:}
\providecommand{\doi}[1]{\href{http://dx.doi.org/#1}{\path{#1}}}
\providecommand{\Pubmed}[1]{\href{pmid:#1}{\path{#1}}}
\providecommand{\bibinfo}[2]{#2}
\ifx\xfnm\relax \def\xfnm[#1]{\unskip,\space#1}\fi
\bibitem[{Bauersfeld and Scaramuzza(2021)}]{POFlight}
\bibinfo{author}{Bauersfeld, L.}, \bibinfo{author}{Scaramuzza, D.},
  \bibinfo{year}{2021}.
\newblock \bibinfo{title}{Range, endurance, and optimal speed estimates for
  multicopters}.
\newblock \bibinfo{journal}{CoRR} \bibinfo{volume}{abs/2109.04741}.
\newblock \URLprefix \url{https://arxiv.org/abs/2109.04741},
  \href{http://arxiv.org/abs/2109.04741}{\tt arXiv:2109.04741}.
\bibitem[{Bicego et~al.(2020)Bicego, Mazzetto, Carli, Farina and
  Franchi}]{Bicego2020}
\bibinfo{author}{Bicego, D.}, \bibinfo{author}{Mazzetto, J.},
  \bibinfo{author}{Carli, R.}, \bibinfo{author}{Farina, M.},
  \bibinfo{author}{Franchi, A.}, \bibinfo{year}{2020}.
\newblock \bibinfo{title}{Nonlinear model predictive control with enhanced
  actuator model for multi-rotor aerial vehicles with generic designs}.
\newblock \bibinfo{journal}{Journal of Intelligent {\&} Robotic Systems}
  \bibinfo{volume}{100}, \bibinfo{pages}{1213--1247}.
\bibitem[{Faessler et~al.(2017)Faessler, Franchi and Scaramuzza}]{Faessler2018}
\bibinfo{author}{Faessler, M.}, \bibinfo{author}{Franchi, A.},
  \bibinfo{author}{Scaramuzza, D.}, \bibinfo{year}{2017}.
\newblock \bibinfo{title}{Differential flatness of quadrotor dynamics subject
  to rotor drag for accurate tracking of high-speed trajectories}.
\newblock \bibinfo{journal}{IEEE Robotics and Automation Letters}
  \bibinfo{volume}{3}, \bibinfo{pages}{620--626}.
\bibitem[{Foehn et~al.(2021)Foehn, Romero and Scaramuzza}]{TimeOptimalPlanning}
\bibinfo{author}{Foehn, P.}, \bibinfo{author}{Romero, A.},
  \bibinfo{author}{Scaramuzza, D.}, \bibinfo{year}{2021}.
\newblock \bibinfo{title}{Time-optimal planning for quadrotor waypoint flight}.
\newblock \bibinfo{journal}{Science Robotics} \bibinfo{volume}{6},
  \bibinfo{pages}{eabh1221}.
\bibitem[{Fourer et~al.(1990)Fourer, Gay and Kernighan}]{fourer1990modeling}
\bibinfo{author}{Fourer, R.}, \bibinfo{author}{Gay, D.M.},
  \bibinfo{author}{Kernighan, B.W.}, \bibinfo{year}{1990}.
\newblock \bibinfo{title}{A modeling language for mathematical programming}.
\newblock \bibinfo{journal}{Management Science} \bibinfo{volume}{36},
  \bibinfo{pages}{519--554}.
\bibitem[{Gati(2013)}]{Gati2013}
\bibinfo{author}{Gati, B.}, \bibinfo{year}{2013}.
\newblock \bibinfo{title}{Open source autopilot for academic research-the
  paparazzi system}, in: \bibinfo{booktitle}{American Control Conference (ACC),
  2013}, \bibinfo{organization}{IEEE}, \bibinfo{address}{Washington, DC}. pp.
  \bibinfo{pages}{1478--1481}.
\newblock \DOIprefix\doi{10.1109/ACC.2013.6580045}.
\bibitem[{Geisert and Mansard(2016)}]{numericaltrajgen}
\bibinfo{author}{Geisert, M.}, \bibinfo{author}{Mansard, N.},
  \bibinfo{year}{2016}.
\newblock \bibinfo{title}{Trajectory generation for quadrotor based systems
  using numerical optimal control}.
\newblock \bibinfo{journal}{CoRR} \bibinfo{volume}{abs/1602.01949}.
\newblock \URLprefix \url{http://arxiv.org/abs/1602.01949},
  \href{http://arxiv.org/abs/1602.01949}{\tt arXiv:1602.01949}.
\bibitem[{Gill et~al.(2005)Gill, Murray and Saunders}]{gill2005snopt}
\bibinfo{author}{Gill, P.E.}, \bibinfo{author}{Murray, W.},
  \bibinfo{author}{Saunders, M.A.}, \bibinfo{year}{2005}.
\newblock \bibinfo{title}{Snopt: An sqp algorithm for large-scale constrained
  optimization}.
\newblock \bibinfo{journal}{SIAM review} \bibinfo{volume}{47},
  \bibinfo{pages}{99--131}.
\bibitem[{Hanover et~al.(2022)Hanover, Foehn, Sun, Kaufmann and
  Scaramuzza}]{AdaptiveNMPC}
\bibinfo{author}{Hanover, D.}, \bibinfo{author}{Foehn, P.},
  \bibinfo{author}{Sun, S.}, \bibinfo{author}{Kaufmann, E.},
  \bibinfo{author}{Scaramuzza, D.}, \bibinfo{year}{2022}.
\newblock \bibinfo{title}{Performance, precision, and payloads: Adaptive
  nonlinear mpc for quadrotors}.
\newblock \bibinfo{journal}{IEEE Robotics and Automation Letters}
  \bibinfo{volume}{7}, \bibinfo{pages}{690--697}.
\bibitem[{Hassanalian and Abdelkefi(2017)}]{Hassanalian2017ClassificationsAA}
\bibinfo{author}{Hassanalian, M.}, \bibinfo{author}{Abdelkefi, A.},
  \bibinfo{year}{2017}.
\newblock \bibinfo{title}{Classifications, applications, and design challenges
  of drones: A review}.
\newblock \bibinfo{journal}{Progress in Aerospace Sciences}
  \bibinfo{volume}{91}, \bibinfo{pages}{99--131}.
\bibitem[{Hwangbo et~al.(2017)Hwangbo, Sa, Siegwart and Hutter}]{RL_wp_stab}
\bibinfo{author}{Hwangbo, J.}, \bibinfo{author}{Sa, I.},
  \bibinfo{author}{Siegwart, R.}, \bibinfo{author}{Hutter, M.},
  \bibinfo{year}{2017}.
\newblock \bibinfo{title}{Control of a quadrotor with reinforcement learning}.
\newblock \bibinfo{journal}{IEEE Robotics and Automation Letters}
  \bibinfo{volume}{2}, \bibinfo{pages}{2096--2103}.
\bibitem[{Kaufmann et~al.(2020)Kaufmann, Loquercio, Ranftl, M{\"{u}}ller,
  Koltun and Scaramuzza}]{deep_drone_acrobatics}
\bibinfo{author}{Kaufmann, E.}, \bibinfo{author}{Loquercio, A.},
  \bibinfo{author}{Ranftl, R.}, \bibinfo{author}{M{\"{u}}ller, M.},
  \bibinfo{author}{Koltun, V.}, \bibinfo{author}{Scaramuzza, D.},
  \bibinfo{year}{2020}.
\newblock \bibinfo{title}{{Deep Drone Acrobatics}}, in:
  \bibinfo{booktitle}{RSS: Robotics, Science, and Systems},
  \bibinfo{publisher}{Robotics: Science and Systems Foundation},
  \bibinfo{address}{Corvalis, Oregon, USA}. pp. \bibinfo{pages}{1--10}.
\newblock \href{http://arxiv.org/abs/2006.05768}{\tt arXiv:2006.05768}.
\bibitem[{Li et~al.(2017)Li, Qian, Zhu, Bao, Helwa and
  Schoellig}]{DNN_improved_tracking}
\bibinfo{author}{Li, Q.}, \bibinfo{author}{Qian, J.}, \bibinfo{author}{Zhu,
  Z.}, \bibinfo{author}{Bao, X.}, \bibinfo{author}{Helwa, M.K.},
  \bibinfo{author}{Schoellig, A.P.}, \bibinfo{year}{2017}.
\newblock \bibinfo{title}{Deep neural networks for improved, impromptu
  trajectory tracking of quadrotors}, in: \bibinfo{booktitle}{2017 IEEE
  International Conference on Robotics and Automation (ICRA)},
  \bibinfo{organization}{IEEE}. pp. \bibinfo{pages}{5183--5189}.
\bibitem[{Li et~al.(2020)Li, \"{O}zt\"{u}rk, {De Wagter}, {de Croon} and
  Izzo}]{DBLP:journals/corr/abs-1912-07067}
\bibinfo{author}{Li, S.}, \bibinfo{author}{\"{O}zt\"{u}rk, E.},
  \bibinfo{author}{{De Wagter}, C.}, \bibinfo{author}{{de Croon}, G.C.H.E.},
  \bibinfo{author}{Izzo, D.}, \bibinfo{year}{2020}.
\newblock \bibinfo{title}{Aggressive online control of a quadrotor via deep
  network representations of optimality principles}, in:
  \bibinfo{booktitle}{2020 IEEE International Conference on Robotics and
  Automation, ICRA 2020}, \bibinfo{publisher}{Institute of Electrical and
  Electronics Engineers (IEEE)}, \bibinfo{address}{United States}. pp.
  \bibinfo{pages}{6282--6287}.
\bibitem[{Li et~al.(2016)Li, Wang, Tan and Zheng}]{RBFNN}
\bibinfo{author}{Li, S.}, \bibinfo{author}{Wang, Y.}, \bibinfo{author}{Tan,
  J.}, \bibinfo{author}{Zheng, Y.}, \bibinfo{year}{2016}.
\newblock \bibinfo{title}{Adaptive rbfnns/integral sliding mode control for a
  quadrotor aircraft}.
\newblock \bibinfo{journal}{Neurocomputing} \bibinfo{volume}{216},
  \bibinfo{pages}{126--134}.
\newblock \URLprefix
  \url{https://www.sciencedirect.com/science/article/pii/S0925231216307780}.
\bibitem[{Liu et~al.(2015)Liu, Lu and Chen}]{explicitMPC}
\bibinfo{author}{Liu, C.}, \bibinfo{author}{Lu, H.}, \bibinfo{author}{Chen,
  W.H.}, \bibinfo{year}{2015}.
\newblock \bibinfo{title}{An explicit mpc for quadrotor trajectory tracking},
  in: \bibinfo{booktitle}{2015 34th Chinese Control Conference (CCC)}, pp.
  \bibinfo{pages}{4055--4060}.
\bibitem[{Mao et~al.(2018)Mao, Szmuk, Xu and
  A{\c{c}}ikmese}]{successive_convexification}
\bibinfo{author}{Mao, Y.}, \bibinfo{author}{Szmuk, M.}, \bibinfo{author}{Xu,
  X.}, \bibinfo{author}{A{\c{c}}ikmese, B.}, \bibinfo{year}{2018}.
\newblock \bibinfo{title}{Successive convexification: A superlinearly
  convergent algorithm for non-convex optimal control problems}.
\newblock \bibinfo{journal}{arXiv preprint arXiv:1804.06539} .
\bibitem[{Mellinger and Kumar(2011)}]{MinSnap}
\bibinfo{author}{Mellinger, D.}, \bibinfo{author}{Kumar, V.},
  \bibinfo{year}{2011}.
\newblock \bibinfo{title}{Minimum snap trajectory generation and control for
  quadrotors}, in: \bibinfo{booktitle}{2011 IEEE International Conference on
  Robotics and Automation}, pp. \bibinfo{pages}{2520--2525}.
\bibitem[{Mueller et~al.(2015)Mueller, Hehn and D'Andrea}]{Mueller}
\bibinfo{author}{Mueller, M.W.}, \bibinfo{author}{Hehn, M.},
  \bibinfo{author}{D'Andrea, R.}, \bibinfo{year}{2015}.
\newblock \bibinfo{title}{A computationally efficient motion primitive for
  quadrocopter trajectory generation}.
\newblock \bibinfo{journal}{IEEE Transactions on Robotics}
  \bibinfo{volume}{31}, \bibinfo{pages}{1294--1310}.
\bibitem[{Romero et~al.(2022)Romero, Penicka and
  Scaramuzza}]{TimeOpimalReplanning}
\bibinfo{author}{Romero, A.}, \bibinfo{author}{Penicka, R.},
  \bibinfo{author}{Scaramuzza, D.}, \bibinfo{year}{2022}.
\newblock \bibinfo{title}{Time-optimal online replanning for agile quadrotor
  flight}.
\newblock \bibinfo{journal}{{IEEE} Robotics and Automation Letters}
  \bibinfo{volume}{7}, \bibinfo{pages}{7730--7737}.
\bibitem[{Romero et~al.(2021)Romero, Sun, Foehn and Scaramuzza}]{MPCC}
\bibinfo{author}{Romero, A.}, \bibinfo{author}{Sun, S.},
  \bibinfo{author}{Foehn, P.}, \bibinfo{author}{Scaramuzza, D.},
  \bibinfo{year}{2021}.
\newblock \bibinfo{title}{Model predictive contouring control for
  near-time-optimal quadrotor flight}.
\newblock \bibinfo{journal}{CoRR} \bibinfo{volume}{abs/2108.13205}.
\newblock \URLprefix \url{https://arxiv.org/abs/2108.13205},
  \href{http://arxiv.org/abs/2108.13205}{\tt arXiv:2108.13205}.
\bibitem[{Ru and Subbarao(2017)}]{aerospace4020031}
\bibinfo{author}{Ru, P.}, \bibinfo{author}{Subbarao, K.}, \bibinfo{year}{2017}.
\newblock \bibinfo{title}{Nonlinear model predictive control for unmanned
  aerial vehicles}.
\newblock \bibinfo{journal}{Aerospace} \bibinfo{volume}{4}.
\newblock \URLprefix \url{https://www.mdpi.com/2226-4310/4/2/31}.
\bibitem[{Smeur et~al.(2016)Smeur, Chu and de~Croon}]{Smeur2016}
\bibinfo{author}{Smeur, E.J.J.}, \bibinfo{author}{Chu, Q.},
  \bibinfo{author}{de~Croon, G.C.H.E.}, \bibinfo{year}{2016}.
\newblock \bibinfo{title}{Adaptive incremental nonlinear dynamic inversion for
  attitude control of micro air vehicles}.
\newblock \bibinfo{journal}{Journal of Guidance, Control, and Dynamics}
  \bibinfo{volume}{39}, \bibinfo{pages}{450--461}.
\bibitem[{Song et~al.(2021)Song, Steinweg, Kaufmann and
  Scaramuzza}]{Autonomous_Drone_Racing_with_Deep_Reinforcement_Learning}
\bibinfo{author}{Song, Y.}, \bibinfo{author}{Steinweg, M.},
  \bibinfo{author}{Kaufmann, E.}, \bibinfo{author}{Scaramuzza, D.},
  \bibinfo{year}{2021}.
\newblock \bibinfo{title}{Autonomous drone racing with deep reinforcement
  learning}, in: \bibinfo{booktitle}{2021 IEEE/RSJ International Conference on
  Intelligent Robots and Systems (IROS)}, \bibinfo{organization}{IEEE}. pp.
  \bibinfo{pages}{1205--1212}.
\bibitem[{Sun et~al.(2022)Sun, Romero, Foehn, Kaufmann and
  Scaramuzza}]{MPC_DFBP}
\bibinfo{author}{Sun, S.}, \bibinfo{author}{Romero, A.},
  \bibinfo{author}{Foehn, P.}, \bibinfo{author}{Kaufmann, E.},
  \bibinfo{author}{Scaramuzza, D.}, \bibinfo{year}{2022}.
\newblock \bibinfo{title}{A comparative study of nonlinear mpc and
  differential-flatness-based control for quadrotor agile flight}.
\newblock \bibinfo{journal}{IEEE Transactions on Robotics} .
\bibitem[{Sun et~al.(2019)Sun, de~Visser and Chu}]{Sun2019}
\bibinfo{author}{Sun, S.}, \bibinfo{author}{de~Visser, C.C.},
  \bibinfo{author}{Chu, Q.}, \bibinfo{year}{2019}.
\newblock \bibinfo{title}{Quadrotor gray-box model identification from
  high-speed flight data}.
\newblock \bibinfo{journal}{Journal of Aircraft} \bibinfo{volume}{56},
  \bibinfo{pages}{645--661}.
\bibitem[{Svacha et~al.(2017)Svacha, Mohta and Kumar}]{Svacha2017ImprovingQT}
\bibinfo{author}{Svacha, J.}, \bibinfo{author}{Mohta, K.},
  \bibinfo{author}{Kumar, V.R.}, \bibinfo{year}{2017}.
\newblock \bibinfo{title}{Improving quadrotor trajectory tracking by
  compensating for aerodynamic effects}.
\newblock \bibinfo{journal}{2017 International Conference on Unmanned Aircraft
  Systems (ICUAS)} , \bibinfo{pages}{860--866}.
\bibitem[{Sánchez-Sánchez and Izzo(2016)}]{LandingProblems}
\bibinfo{author}{Sánchez-Sánchez, C.}, \bibinfo{author}{Izzo, D.},
  \bibinfo{year}{2016}.
\newblock \bibinfo{title}{Real-time optimal control via deep neural networks:
  Study on landing problems}.
\newblock \bibinfo{journal}{Journal of Guidance, Control, and Dynamics}
  \bibinfo{volume}{41}.
\bibitem[{Tailor and Izzo(2019)}]{Tailor}
\bibinfo{author}{Tailor, D.}, \bibinfo{author}{Izzo, D.}, \bibinfo{year}{2019}.
\newblock \bibinfo{title}{Learning the optimal state-feedback via supervised
  imitation learning}.
\newblock \bibinfo{journal}{Astrodynamics} \bibinfo{volume}{3},
  \bibinfo{pages}{361--374}.
\bibitem[{Tal and Karaman(2020)}]{tal2020accurate}
\bibinfo{author}{Tal, E.}, \bibinfo{author}{Karaman, S.}, \bibinfo{year}{2020}.
\newblock \bibinfo{title}{Accurate tracking of aggressive quadrotor
  trajectories using incremental nonlinear dynamic inversion and differential
  flatness}.
\newblock \bibinfo{journal}{IEEE Transactions on Control Systems Technology}
  \bibinfo{volume}{29}, \bibinfo{pages}{1203--1218}.
\bibitem[{Tang et~al.(2018)Tang, Sun and Hauser}]{LearnTraj}
\bibinfo{author}{Tang, G.}, \bibinfo{author}{Sun, W.}, \bibinfo{author}{Hauser,
  K.}, \bibinfo{year}{2018}.
\newblock \bibinfo{title}{Learning trajectories for real- time optimal control
  of quadrotors}.
\newblock \bibinfo{journal}{IEEE/RSJ Intl Conf on Intelligent Robots and
  Systems} \URLprefix \url{https://par.nsf.gov/biblio/10100589}.
\bibitem[{Tankasala et~al.(2022)Tankasala, Pehlivanturk, Bakolas and
  Pryor}]{tankasala2022smooth}
\bibinfo{author}{Tankasala, S.}, \bibinfo{author}{Pehlivanturk, C.},
  \bibinfo{author}{Bakolas, E.}, \bibinfo{author}{Pryor, M.},
  \bibinfo{year}{2022}.
\newblock \bibinfo{title}{Smooth time optimal trajectory generation for
  drones}.
\newblock \bibinfo{journal}{arXiv preprint arXiv:2202.09392} .
\bibitem[{Torrente et~al.(2021)Torrente, Kaufmann, Foehn and
  Scaramuzza}]{torrente2021data}
\bibinfo{author}{Torrente, G.}, \bibinfo{author}{Kaufmann, E.},
  \bibinfo{author}{Foehn, P.}, \bibinfo{author}{Scaramuzza, D.},
  \bibinfo{year}{2021}.
\newblock \bibinfo{title}{Data-driven {MPC} for quadrotors}.
\newblock \bibinfo{journal}{{IEEE} Robotics and Automation Letters}
  \bibinfo{volume}{6}, \bibinfo{pages}{3769--3776}.
\newblock \DOIprefix\doi{10.1109/lra.2021.3061307},
  \href{http://arxiv.org/abs/2102.05773}{\tt arXiv:2102.05773}.
\bibitem[{Van~Nieuwstadt and Murray(1998)}]{NIEUWSTADT19962301}
\bibinfo{author}{Van~Nieuwstadt, M.J.}, \bibinfo{author}{Murray, R.M.},
  \bibinfo{year}{1998}.
\newblock \bibinfo{title}{Real-time trajectory generation for differentially
  flat systems}.
\newblock \bibinfo{journal}{International Journal of Robust and Nonlinear
  Control: IFAC-Affiliated Journal} \bibinfo{volume}{8},
  \bibinfo{pages}{995--1020}.
\bibitem[{Yu et~al.(2022)Yu, Nagpal, Mceowen, A{\c{c}}{\i}kme{\c{s}}e and
  Topcu}]{yu2022real}
\bibinfo{author}{Yu, Y.}, \bibinfo{author}{Nagpal, K.},
  \bibinfo{author}{Mceowen, S.}, \bibinfo{author}{A{\c{c}}{\i}kme{\c{s}}e, B.},
  \bibinfo{author}{Topcu, U.}, \bibinfo{year}{2022}.
\newblock \bibinfo{title}{Real-time quadrotor trajectory optimization with
  time-triggered corridor constraints}.
\newblock \bibinfo{journal}{arXiv preprint arXiv:2208.07259} .

\end{thebibliography}

\bio{figures/robin}
\textbf{Robin Ferede} received the M.Sc. degree in aerospace engineering from Delft University of Technology, Delft, The Netherlands, in 2022. His graduation work focused on end-to-end neural network-based optimal quadcopter control. Since 2022, he has been working toward a Ph.D. degree. His research interests lie in combining optimal control theory with machine learning algorithms to address the challenges associated with autonomous quadcopter flight.
\endbio

\bio{figures/guido}
\textbf{Guido de Croon} received his M.Sc. and Ph.D. in
the field of Artificial Intelligence (AI) at Maastricht
University, the Netherlands. His research interest lies
in computationally efficient algorithms for robot
autonomy, with an emphasis on computer vision and
evolutionary robotics. Since 2008 he has worked on
algorithms for achieving autonomous flight with small
and lightweight flying robots, such as the DelFly flapping wing MAV. In 2011-2012, he was a research fellow
in the Advanced Concepts Team of the European Space
Agency, where he studied topics such as optical flow-based control algorithms for extraterrestrial landing scenarios. Currently, he is
a full professor at TU Delft and scientific lead of the Micro Air Vehicle lab
(MAV-lab) of the Delft University of Technology.
\endbio

\bio{figures/christophe}
\textbf{Christophe De Wagter} received his M.Sc. in Aerospace Engineering at the Delft University of Technology in 2004 on the topic of vision-based control. In 2005 he created the \href{http://mavlab.tudelft.nl}{Micro Air Vehicle Lab} where he worked as a researcher until he obtained a Ph.D. in robotics. His areas of interest range from control theory and sensor fusion to
computer vision, electronics and AI. He proposed novel concepts like the \href{http://www.delfly.nl}{DelFly}, and worked on the \href{http://www.delftacopter.nl}{DelftaCopter} and hydrogen-powered \href{http://www.nederdrone.nl}{Nederdrone}.
In parallel, he has been a freelance electronics and software developer for local startup companies and is a private pilot, glider pilot, and certified drone pilot. He is also the safety manager for the MAVLab drone operations.
Over the years he won many awards ranging from the 1st prize for ``Best Fully Autonomous Indoor MAV'' at the EMAV 2008 in Braunschweig, to the ``World Champion in Artificial Intelligence Drone Racing,'' at the AIRR-2019 in the United States.
\endbio

\bio{figures/dario}
\textbf{Dario Izzo} graduated as a Doctor of Aeronautical Engineering from the University Sapienza of Rome (Italy). He then took a second master in Satellite Platforms at the University of Cranfield in the United Kingdom and completed his Ph.D. in Mathematical Modelling at the University Sapienza of Rome where he lectured classical mechanics and space flight mechanics. Dario Izzo later joined the European Space Agency and became the scientific coordinator of its Advanced Concepts Team. He devised and managed the Global Trajectory Optimization Competitions events, the ESA Summer of Code in Space and the Kelvins innovation and competition platform. He published more than 170 papers in international journals and conferences making key contributions to the understanding of flight mechanics and spacecraft control and pioneering techniques based on evolutionary and machine-learning approaches. Dario Izzo received the Humies Gold Medal and led the team winning the 8th edition of the Global Trajectory Optimization Competition
\endbio

\end{document}